\documentclass[10pt,twocolumn,letterpaper]{article}
\usepackage[pagenumbers]{cvpr}
\usepackage{times,graphicx,amsmath,amssymb,booktabs,multirow,xcolor,placeins}
\definecolor{cvprblue}{rgb}{0.21,0.49,0.74}
\usepackage{makecell,threeparttable}
\usepackage[breaklinks,colorlinks,allcolors=cvprblue]{hyperref}
\newcommand{\method}{C\textsuperscript{2}-Routing}

\newcommand{\E}{\mathbb{E}}

\title{\method: Complementary Cropping and Compression\\for Efficient Vision-Language Inference}
\author{Guorui Lu \qquad Qinyu Chen\\
Leiden University\\
{\tt\small g.lu@liacs.leidenuniv.nl \quad q.chen@liacs.leidenuniv.nl}}

\hypersetup{
  pdftitle={C2-Routing: Complementary Cropping and Compression for Efficient Vision-Language Inference},
  pdfauthor={Guorui Lu and Qinyu Chen},
  pdfkeywords={Vision-language models, Efficient inference, Visual token reduction, Model-agnostic input routing, Adaptive cropping and compression}
}
\begin{document}
\maketitle

\begin{abstract}
Visual Question Answering (VQA) imposes substantial computational overhead on vision-language models (VLMs), motivating model-agnostic input reduction methods that can be readily applied across diverse VLMs, including in black-box settings. Simply enforcing a visual-token cap reduces computation, but it can discard information needed to answer the question. A single reduction strategy also struggles to accommodate different scenarios. Our key observation is that cropping and compression are complementary: cropping highlights small targets and local details, whereas whole-image compression better preserves context for large targets or distributed evidence. Based on this observation, we propose \method, which adaptively selects cropping or compression according to the input image and question. Since both strategies may still discard key information, we further design a safety gate that uses the predicted joint correctness distribution under original and reduced inputs to guide fallback and protect accuracy. To accommodate consumer GPU resource constraints, all training and evaluation experiments run on RTX 3090 GPUs. To measure gains beyond token capping, we use full-image baselines with the same cap of 1,024 visual tokens. We conduct extensive evaluation on three VLMs and 18 datasets spanning local details, global scenes, scene text, and documents. With the visual budget already capped, \method\ further reduces end-to-end FLOPs by 36.53\% and visual tokens by 42.36\% on average, with a mean relative task-score decrease of only 0.77\%.
\end{abstract}

\section{Introduction}
\begin{figure*}[t]
\centering\includegraphics[width=\textwidth]{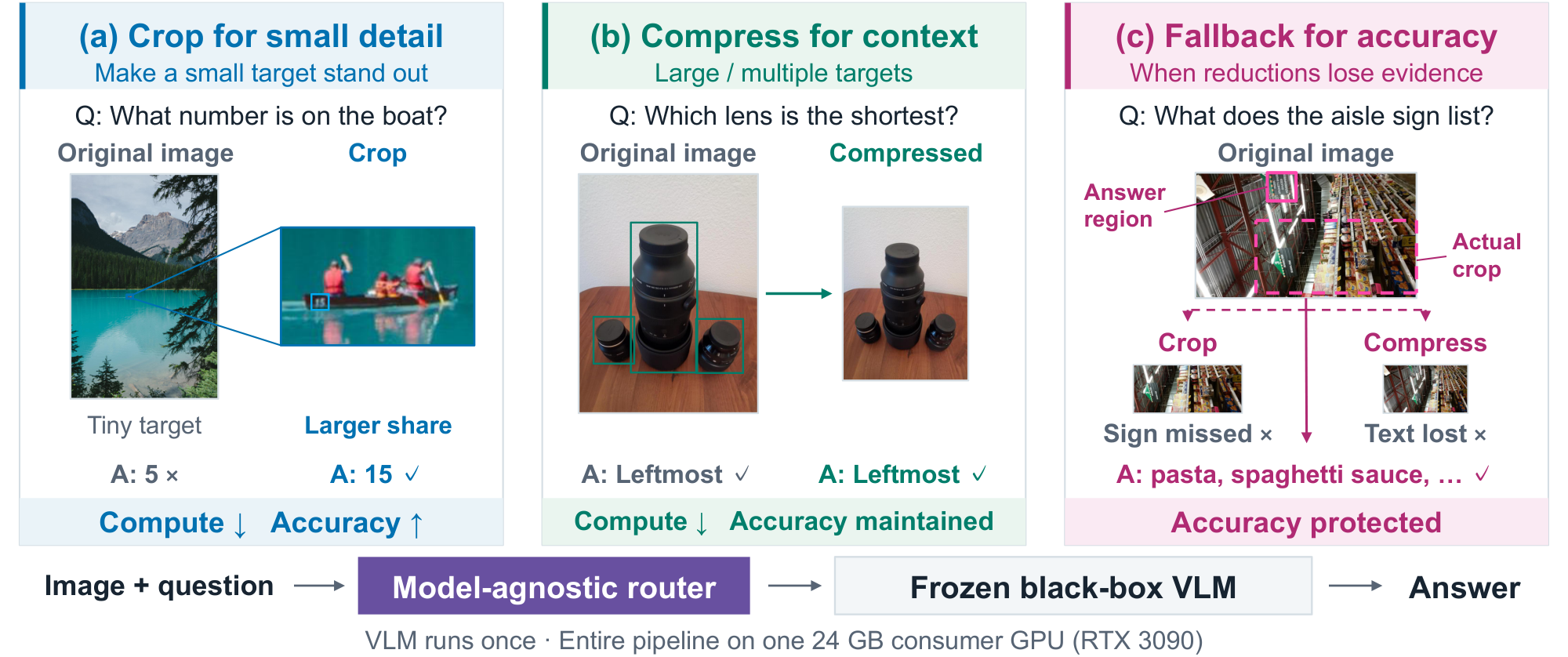}
\caption{\textbf{Complementary cropping and compression with fallback protection.} (a) For targets occupying a small fraction of the full image, cropping highlights the key region, reducing computation and improving accuracy. (b) For large targets or key information distributed across multiple locations, compression preserves global context while reducing computation and maintaining accuracy. (c) When cropping or compression may discard key information, fallback to the original input protects accuracy.}
\label{fig:overview}
\end{figure*}

Vision-language models (VLMs) have made significant progress in visual understanding and multimodal reasoning~\cite{llavaonevision,qwen25vl,internvl3}. However, the large number of visual tokens generated from images greatly increases computation and memory requirements during inference~\cite{cares}, raising deployment and usage costs. In the single-image question-answering example in Fig.~\ref{fig:overview}(a), Qwen2.5-VL-7B~\cite{qwen25vl} encodes an image preprocessed to $700 \times 1092$ pixels into 975 visual tokens, accounting for 94.1\% of the 1,036 input tokens. Reducing visual input while preserving task-relevant information is therefore important for lowering VLM deployment and usage costs.

Existing studies reduce visual computation primarily through internal visual token reduction or input image processing. Visual token pruning and merging typically rely on model internals~\cite{fastv,visionzip,sparsevlm,pyramiddrop}, complicating adaptation across models and limiting their use in black-box settings. Input-level methods process images through question-relevant cropping~\cite{cropvlm}, dynamic whole-image resolution adjustment~\cite{cares,inputadaptive}, or spatially varying resampling~\cite{llmind}. However, cropping may discard necessary context, whereas compression may damage key details, limiting the applicability of either strategy alone across scenarios.

A key finding of this study is that cropping and compression are suited to complementary scenarios, as illustrated in Fig.~\ref{fig:overview}(a,b). When question-relevant information is concentrated in a small region, whole-image downsampling leaves fewer pixels to represent that region, potentially making answer-relevant details unrecognizable. Cropping removes irrelevant background and increases the key region's relative prominence, helping the model attend to and recognize the evidence. It can therefore reduce visual tokens and computation while improving accuracy. Conversely, when relevant information spans a large area or is distributed across multiple locations, a crop that retains all relevant content typically removes little background and offers limited computational savings; a tighter crop can discard necessary evidence and context. Moderate whole-image downsampling can instead preserve recognizable evidence and spatial relationships, reducing visual tokens and computation while largely maintaining task performance.

Furthermore, as shown in Fig.~\ref{fig:overview}(c), both cropping and compression can remove information needed to answer the question, degrading performance. The input-processing pipeline therefore requires a protection mechanism that falls back to the original input when the predicted risk of performance loss is high. Compared with direct original-input inference, fallback requests incur the extra cost of the decision and preprocessing stages already executed. Nevertheless, the complete pipeline can still reduce overall computation if inference savings from requests without fallback exceed frontend overhead across all requests. Prior work in an EEG-guided setting has shown that selective cropping with fallback to the original input can largely preserve or improve accuracy while reducing overall computation, even after accounting for all stages~\cite{brainfocus}.

Based on these observations, we propose \method, a visual input router that combines complementary reduction strategies with accuracy protection. Its core design separates strategy selection from risk assessment. The former chooses cropping or whole-image compression based on the image and question, exploiting their strengths for different spatial distributions of task-relevant information. The latter estimates the chosen reduction's performance impact relative to the original input to decide whether to retain it or fall back. To assess how input reduction affects accuracy, we model the joint distribution of answer correctness under the original and reduced inputs. Strategy selection and risk prediction share an image--question representation to reduce frontend overhead. All routing and fallback decisions precede target inference and require no access to target-model internals, enabling black-box use.

\begin{figure*}[t]
\centering\includegraphics[width=\textwidth]{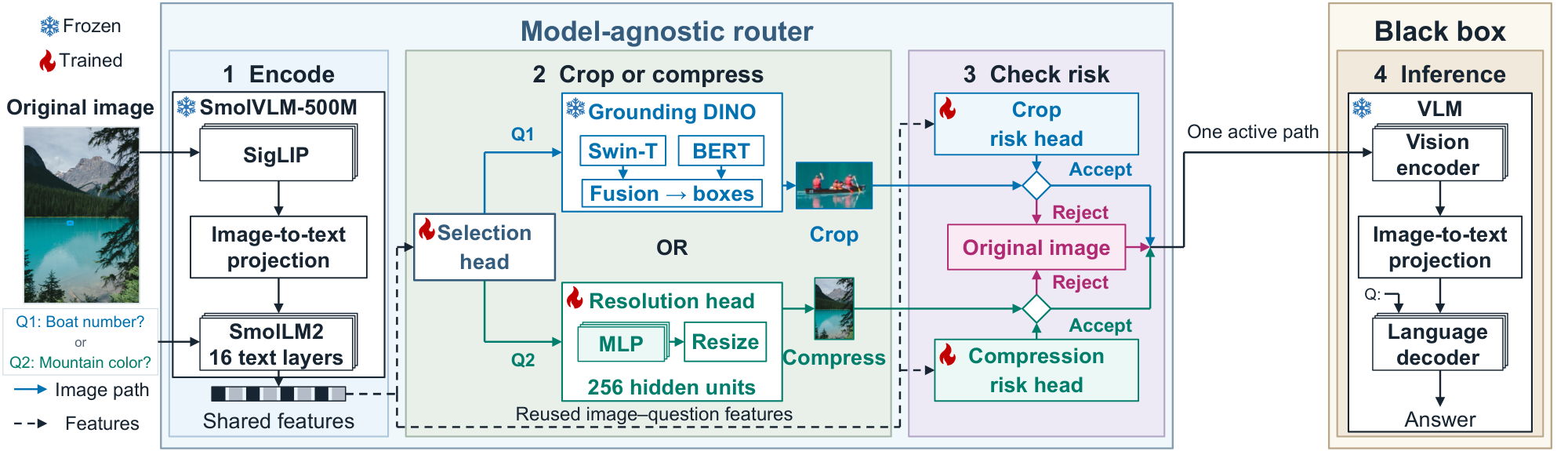}
\caption{\textbf{Overview of C$^{2}$-Routing.} Shared image--question features guide cropping or compression. A risk head predicts joint answer correctness under original and reduced inputs, allowing the safety gate to accept the reduction or fall back to the original image. The VLM answers from the resulting image and question.}
\label{fig:method}
\end{figure*}

Our full-image budget comparison shows that a visual-token cap alone considerably reduces computation at an accuracy cost. This observation highlights the role of visual budgets in efficiency evaluation. We therefore compare methods against full-image baselines with the same cap of 1,024 visual tokens to measure the additional savings and accuracy changes due to input routing. We jointly control visual budgets and frontend costs to meet consumer GPU resource constraints, running all training and evaluation experiments on RTX 3090 GPUs. Experiments on three VLMs and 18 datasets show that our method further reduces end-to-end computation while protecting accuracy under an already restricted visual budget.

Our main contributions are:
\begin{itemize}
\item \textbf{Complementarity-driven input routing.} We identify the complementary strengths of cropping and compression for local-detail and global-context tasks, and propose \method\ to select the strategy without accessing target-model internals, enabling black-box use.
\item \textbf{Accuracy protection through joint correctness prediction.} We jointly model answer correctness under original and reduced inputs to distinguish harmful from beneficial reductions, guiding fallback to protect accuracy.
\item \textbf{Visual budgets in efficiency evaluation.} Full-image budget comparisons reveal significant computational savings and an accuracy cost from token capping alone. Comparisons against full-image baselines with the same cap quantify the additional gains from input routing.
\item \textbf{Broad evaluation under consumer GPU constraints.} All training and evaluation experiments run on RTX 3090 GPUs. Experiments on three VLMs and 18 datasets show that, even when the visual-token budget is already capped, our method further reduces computation after accounting for all frontend overhead while largely preserving accuracy, demonstrating its effectiveness and broad applicability under resource constraints.
\end{itemize}

\section{\texorpdfstring{Related Work}{Related Work}}
\subsection{\texorpdfstring{Visual Token Pruning and Merging}{Visual Token Pruning and Merging}}
Visual token pruning and merging use internal attention or hidden representations to remove or merge redundant tokens at the vision encoder output or within language-model layers~\cite{fastv,visionzip,sparsevlm,pyramiddrop}. For example, VisionZip~\cite{visionzip} and HiPrune~\cite{hiprune} both use attention information from the vision encoder to reduce the number of visual tokens passed to the language model. Such changes require access to model internals, complicating adaptation across models and preventing application in settings where only input--output interfaces are available, for example, GPT-4V APIs~\cite{gpt4,gpt4v}. \method\ instead crops or resizes input images before inference, avoiding model-specific modifications.

\subsection{\texorpdfstring{Input-Level Image Processing}{Input-Level Image Processing}}
\noindent\textbf{Cropping and visual search.}\quad
Cropping and visual search improve fine-detail recognition by highlighting question-relevant local information, but existing approaches have limitations in model independence and inference cost. ViCrop~\cite{vicrop} and GapSight~\cite{gapsight} require access to target-model internal states, while AdaptVision~\cite{adaptvision} requires training the target model, limiting their adaptation across models and applicability to black-box settings. ViCrop~\cite{vicrop} and ZoomEye~\cite{zoomeye} improve accuracy through visual cropping and iterative region exploration, respectively, at the cost of additional computation compared with direct full-image inference. CropVLM~\cite{cropvlm} combines a low-resolution full image with a learned crop to avoid encoding the full image at high resolution. Moreover, when the target object occupies a large fraction of the image or multiple relevant targets are spatially dispersed, cropping alone often offers limited computational savings and may discard key information needed to answer the question.

\noindent\textbf{Dynamic compression.}\quad
Dynamic compression reduces visual redundancy while preserving global context by selecting whole-image resolution or adjusting spatial sampling density, but limitations remain in deployment flexibility and inference cost. VisionThink~\cite{visionthink} uses reinforcement learning to train the target model to request higher resolution as needed, limiting its direct applicability to frozen black-box models. CARES~\cite{cares} and input-adaptive preprocessing~\cite{inputadaptive} select input resolution before target-model inference using low-resolution proxy features and image-complexity measures, respectively, but may underestimate local detail requirements. Spatially non-uniform sampling methods like ~\cite{seeingmore} and ~\cite{llmind} interpolate sampled images back to their original resolution; the latter also incurs extra inference cost through repeated target-model calls during iterative optimization for each image. Even when compression preserves details in key regions, it may not increase their proportion of the full image, leaving small targets difficult for the model to notice.

\method\ dynamically combines cropping and compression within an input-routing framework, exploiting their complementarity in local-detail and global-context scenarios to address a broader range of visual tasks. We predict the joint distribution of answer correctness under original and reduced inputs to guide fallback and protect accuracy. The method requires neither modifying the VLM nor accessing its internals, making it model-agnostic and easy to apply across models.

\section{Method}
\label{sec:method}
\subsection{Overview}
\label{sec:overview}

Fig.~\ref{fig:method} presents \method, a model-agnostic frontend that selects and protects visual information before VLM inference. A frozen lightweight encoder first extracts shared features from the image and question. A selection head chooses cropping to emphasize local details or whole-image compression to preserve broader context with fewer visual tokens. The crop branch localizes a question-relevant region; the compression branch uses a resolution head to select a token budget for image compression.

The active branch's risk head predicts joint answer correctness under the original and proposed reduced inputs, using frontend features. A safety gate uses this prediction to accept the reduction or fall back to the original input. The resulting image and original question are then passed to the frozen target VLM. Thus adaptive selection exploits crop--compression complementarity, while the safety gate protects accuracy when the proposed reduction loses essential information. All input decisions use frontend features, and the target VLM is called once with the resulting image.

\subsection{Adaptive Cropping and Compression}
\label{sec:adaptive-reduction}
Whether to crop or compress an image depends on its content and the visual information required by the question. Even for the same image, reading small text in a local region may favor cropping, whereas reasoning about relationships among multiple objects may require broader context and favor compression. The router must therefore analyze the image and question together to identify relevant visual information and choose a reduction that preserves it. To balance this capability with routing overhead, a small VLM extracts joint image--question features shared by subsequent decision modules.

\noindent\textbf{Shared image--question features.}\quad
For an image--question pair $x_i=(I_i,Q_i)$, we use frozen SmolVLM-500M~\cite{smolvlm} as the frontend encoder $E$ and follow the feature extraction of CARES~\cite{cares} to obtain $h_i=E(I_i,Q_i)$, the shared image--question features. The selection and resolution heads reuse this representation to consider image content and question requirements without repeated encoding. Both heads use lightweight classifiers to limit decision overhead.

\noindent\textbf{Strategy Selection.}\quad
We separate input reduction into strategy selection and compression-budget selection: the former determines whether to emphasize a local region or preserve whole-image context, while the latter determines how much visual detail to retain when keeping the whole image. The selection head first chooses cropping or compression from $h_i$. On the crop branch, frozen Grounding DINO-tiny~\cite{groundingdino} localizes a question-relevant region, and cropping increases its relative prominence in the input. On the compression branch, the resolution head selects a visual-token cap of 256 or 512 and adjusts the whole-image resolution accordingly. 

Let $F$ denote the original input, $R$ the cropped input, and $C_b$ the compressed whole-image input with a visual-token budget of $b$. Let $\mathcal A_{\rm red}$ denote the set of reduced-input choices. For the $i$-th image--question pair $x_i$, let $a_i^0$ denote the reduced input proposed before the safety gate, where superscript $0$ marks the initial choice. The routing policy $\pi$ produces the final input choice $\pi_i=\pi(x_i)$ for the target VLM. Thus,
\begin{equation}
a_i^0\in\mathcal A_{\rm red}=\{R,C_{256},C_{512}\},
\qquad \pi_i\in\{a_i^0,F\}.
\end{equation}
The safety gate in Sec.~\ref{sec:joint-correctness} accepts $a_i^0$ or falls back to the original input $F$. In the main setting, the original and cropped inputs both use a cap of 1,024 visual tokens; actual token counts are recorded from the target model's encoding of each input.

\noindent\textbf{Generating training labels.}\quad
Before router training, we run each frozen target VLM separately on the original image, crop, and each compressed version with the same question to generate labels. We compare the generated answers with the reference answers under the dataset's evaluation rule to obtain an answer score $S_i^a\in[0,1]$ for each sample and action. For datasets with binary scores, the correctness label $Y_i^a\in\{0,1\}$ equals $S_i^a$; for datasets that award partial credit, we obtain $Y_i^a$ using predefined score thresholds. We use answer scores to generate labels for the resolution head, and correctness labels for strategy selection and joint correctness prediction.

\noindent\textbf{Selection labels and training.}\quad
The selection head chooses cropping or compression based on the image and question to exploit their complementarity. Training prioritizes correct answers and, when correctness is equal, lower computational cost. For a training sample, let $a_i^C\in\{C_{256},C_{512}\}$ be the compression candidate. We compare $Y_i^R$ with $Y_i^{a_i^C}$: if only one candidate answers correctly, its strategy becomes the selection label; when correctness is equal, we choose the lower-cost candidate, including localization and final-answer inference in the crop cost. The selection head learns this decision with weighted cross-entropy. The resolution head is trained first to select compression budgets for generating selection labels. We generate selection labels using the deployment compression rule, so training compares the same input alternatives.

\noindent\textbf{Resolution labels and training.}\quad
Following the minimum-sufficient-resolution rule of CARES~\cite{cares}, a budget is sufficient when its answer score meets the required threshold and further increasing the budget yields little improvement. The training label is the smallest sufficient budget, and the resolution head is trained with cross-entropy.

\subsection{\texorpdfstring{Safety Gate for Accuracy Protection}{Safety Gate for Accuracy Protection}}
\label{sec:joint-correctness}
Because cropping and compression can discard answer-relevant information, we design a safety gate to decide whether to fall back to the original input. Predicting reduced-input errors alone is insufficient: fallback cannot correct them if the original input also fails. Conversely, reduction may highlight key regions or reduce irrelevant distractions, yielding a correct answer where the original input fails; fallback would then make the answer incorrect. We therefore jointly model answer correctness under the original and reduced inputs to distinguish harm that fallback can correct from benefits it may forfeit.

\noindent\textbf{Joint correctness states.}\quad
For sample $i$ and reduced input $a\in\mathcal A_{\rm red}$, $Y_i^F$ and $Y_i^a$ indicate answer correctness under the original and reduced inputs, respectively: 1 for correct and 0 otherwise. Define their joint state $s_i^a$ in the order $(Y_i^F,Y_i^a)$:
\begin{equation}
11\leftrightarrow(1,1),\quad H\leftrightarrow(1,0),\quad
B\leftrightarrow(0,1),\quad 00\leftrightarrow(0,0).
\label{eq:paired-states}
\end{equation}
Here $H$ denotes harm from reduction, $B$ denotes benefit, and $11$ and $00$ denote shared success and shared failure. Fallback corrects errors in $H$ but loses the benefits in $B$, while leaving correctness unchanged in the other two states.

\noindent\textbf{Risk-head inputs and prediction.}\quad
Cropping changes which image regions are retained, whereas compression changes how much detail is preserved across the whole image. The two risk heads therefore use different information about the proposed reduction. Let $z_i^a$ denote the risk-head input features. The crop risk head combines the shared image--question features $h_i$ with geometric and confidence features computed from Grounding DINO's localization results, such as crop area fraction, the number of detected boxes, and localization confidence. The compression risk head combines $h_i$ with the selected compression level to assess its effect on answer correctness. These inputs are available from preceding stages, avoiding an additional encoding of the reduced image for risk assessment.

Let $\mathcal S=\{11,H,B,00\}$ and define the conditional state probabilities as $p_s^a(z_i^a)=\Pr(s_i^a=s\mid z_i^a)$ for $s\in\mathcal S$. The risk head $f_\theta^a$, with trainable parameters $\theta$, produces four state scores from $z_i^a$. Softmax converts these scores into predicted probabilities:
\begin{equation}
\widehat{\mathbf p}^{a}(z_i^a)
=\operatorname{softmax}(f_\theta^a(z_i^a)).
\label{eq:joint-prediction}
\end{equation}
The four components estimate $p_{11}^a(z_i^a)$, $p_H^a(z_i^a)$, $p_B^a(z_i^a)$, and $p_{00}^a(z_i^a)$, respectively. Retaining the full probability distribution allows the safety gate to consider the likelihoods of both harm and benefit.

\noindent\textbf{Estimating the accuracy change.}\quad
$H$ contributes $+1$ to accuracy loss, $B$ contributes $-1$, and the other two states contribute zero. We therefore define the risk score as the expected accuracy loss of the reduced input relative to the original:
\begin{equation}
\begin{aligned}
r_a(z_i^a)&=\E[Y_i^F-Y_i^a\mid z_i^a]\\
&=p_H^a(z_i^a)-p_B^a(z_i^a).
\end{aligned}
\label{eq:risk}
\end{equation}
The reduced-input error probability, $p_H^a(z_i^a)+p_{00}^a(z_i^a)$, includes shared failures that fallback cannot correct. The risk score instead measures the correctness change caused by reducing the input and credits potential benefits. The safety gate uses its prediction, $\widehat r_a(z_i^a)=\widehat p_H^a(z_i^a)-\widehat p_B^a(z_i^a)$, to make decisions.

\noindent\textbf{Deciding whether to fall back.}\quad
The safety gate considers both the expected accuracy change and computational saving when deciding whether to accept the proposed reduced input $a=a_i^0$. Let $q_{i,a}=\Pr(Y_i^a=1\mid z_i^a)$ and $q_{i,F}=\Pr(Y_i^F=1\mid z_i^a)$ denote the probabilities of answering correctly under the reduced and original inputs, respectively. Let $D$ be the number of datasets and $\mathcal V_d$ the validation sample indices for dataset $d$. Let $T_i^a$ and $T_i^F$ denote final-answer inference FLOPs for sample $i$ under the reduced and original inputs, respectively. The mean relative inference cost is
\begin{equation}
\bar\rho_a=\frac{1}{D}\sum_{d=1}^{D}
\frac{\sum_{i\in\mathcal V_d}T_i^a}
{\sum_{i\in\mathcal V_d}T_i^F}.
\label{eq:relative-inference-cost}
\end{equation}

Normalizing original-input inference cost to 1 and using $\lambda\ge0$ to control the accuracy--computation tradeoff, we define utilities $U_{i,a}=q_{i,a}-\lambda\bar\rho_a$ and $U_{i,F}=q_{i,F}-\lambda$. Their difference is
\begin{equation}
U_{i,a}-U_{i,F}=-r_a(z_i^a)+\lambda(1-\bar\rho_a).
\end{equation}
The gate thus accepts the reduced input when the predicted accuracy loss does not exceed the weighted computational saving, and falls back otherwise:
\begin{equation}
\pi(x_i)=
\begin{cases}
a,&\widehat r_a(z_i^a)\le\lambda(1-\bar\rho_a),\\
F,&\text{otherwise}.
\end{cases}
\label{eq:gate}
\end{equation}
The risk score $\widehat r_a$ is computed from the risk head's joint correctness probabilities after calibration on the validation set. The cost ratio $\bar\rho_a$ is estimated from validation-set inference FLOPs and remains fixed during deployment. We jointly select $\lambda$ and the selection and resolution thresholds on the validation set, prioritizing task score among candidates with similar end-to-end FLOPs savings that meet preset performance-loss limits. Each target model uses the selected parameters across all test datasets; search ranges and selection rules are provided in Appendix~\ref{sec:s-head-training}.

\noindent\textbf{Label generation.}\quad
We reuse the offline answers under original and reduced inputs to generate four-state labels for each sample, directly capturing the correctness change caused by input reduction. Because models may respond differently to reduction, we generate each target's labels from its own answers.

\section{Experiments}
\newcommand{\resultblank}{\underline{\hspace{0.55cm}}}
\newcommand{\efficiencycomparisonrows}{%
CARES-F1024~\cite{cares} & 65.65 & 97.71 & -1.373 & 35.69 & 39.36 & 64.33 & 96.46 & -2.265 & 33.03 & 37.12 & 65.81 & 98.87 & -1.077 & 29.89 & 34.98 \\
CropVLM-F1024~\cite{cropvlm} & 65.41 & 97.40 & -1.618 & -35.93 & -8.66 & 64.99 & 97.82 & -1.609 & -56.23 & -8.66 & 65.72 & 98.55 & -1.171 & -49.68 & -21.70\\
}
\newcommand{\accuracyreferencerows}{%
ViCrop 256~\cite{vicrop} & 63.47 & 94.36 & -3.556 & -20.65 & 56.33 & 62.56 & 94.69 & -4.038 & -17.47 & 56.31 & \multicolumn{5}{c}{N/A}\\
ViCrop 1024 & 67.99 & 102.21 & +0.967 & -203.10 & -5.40 & 67.08 & 101.36 & +0.487 & -203.26 & -5.20 & \multicolumn{5}{c}{N/A}\\
ZoomEye~\cite{zoomeye} & 69.09 & 105.55 & +2.066 & -2328.90 & 9.68 & 68.80 & 106.91 & +2.202 & -4603.66 & 16.35 & 70.12 & 108.63 & +3.227 & -5278.93 & 0.95 \\
}

\begin{table*}[tp]
\caption{\textbf{Performance and efficiency averaged over 18 datasets.} Ours: four-seed means. Changes and savings are relative to each model's $F_{1024}$ baseline.}
\label{tab:main}\label{tab:method-comparison}
\centering
\fontsize{7.5}{8.5}\selectfont
\setlength{\tabcolsep}{0.7pt}
\renewcommand{\arraystretch}{1.0}
\begin{threeparttable}
\begin{tabular*}{\textwidth}{@{\extracolsep{\fill}}cccccc@{\hspace{6pt}}ccccc@{\hspace{6pt}}ccccc@{}}\toprule
 & \multicolumn{5}{c}{\textbf{\boldmath Qwen2.5-VL-7B}} & \multicolumn{5}{c}{\textbf{\boldmath Qwen2.5-VL-3B}} & \multicolumn{5}{c}{\textbf{\boldmath InternVL3-8B}}\\
\cmidrule(lr){2-6}\cmidrule(lr){7-11}\cmidrule(lr){12-16}
\textbf{\boldmath Method} & \textbf{\boldmath Score} & \textbf{\boldmath \makecell[cc]{Retention\\(\%)}} & \textbf{\boldmath \makecell[cc]{$\Delta$\\(pp)}} & \textbf{\boldmath \makecell[cc]{FLOPs\\saved (\%)}} & \textbf{\boldmath \makecell[cc]{Tokens\\saved (\%)}} & \textbf{\boldmath Score} & \textbf{\boldmath \makecell[cc]{Retention\\(\%)}} & \textbf{\boldmath \makecell[cc]{$\Delta$\\(pp)}} & \textbf{\boldmath \makecell[cc]{FLOPs\\saved (\%)}} & \textbf{\boldmath \makecell[cc]{Tokens\\saved (\%)}} & \textbf{\boldmath Score} & \textbf{\boldmath \makecell[cc]{Retention\\(\%)}} & \textbf{\boldmath \makecell[cc]{$\Delta$\\(pp)}} & \textbf{\boldmath \makecell[cc]{FLOPs\\saved (\%)}} & \textbf{\boldmath \makecell[cc]{Tokens\\saved (\%)}}\\\midrule
$F_{1024}$ & 67.03 & 100.00 & +0.000 & 0.00 & 0.00 & 66.60 & 100.00 & +0.000 & 0.00 & 0.00 & 66.89 & 100.00 & +0.000 & 0.00 & 0.00\\
\multicolumn{16}{c}{\textbf{Accuracy-oriented references}}\\
\accuracyreferencerows
\midrule
\multicolumn{16}{c}{\textbf{Efficiency-oriented methods}}\\
\efficiencycomparisonrows
Ours Safe-M & 66.79 & 99.44 & -0.234 & 25.35 & 30.42 & 66.70 & 100.19 & +0.099 & 21.52 & 28.92 & 67.75 & 101.44 & +0.864 & 25.38 & 30.84\\
Ours Safe-A & 66.84 & 99.48 & -0.189 & 25.32 & 30.52 & 66.59 & 99.93 & -0.002 & 19.35 & 26.03 & 67.74 & 101.41 & +0.849 & 25.26 & 30.73\\
Ours Balanced-M & 66.11 & 98.40 & -0.911 & 37.39 & 42.34 & 65.71 & 98.42 & -0.887 & 35.21 & 41.13 & 66.89 & 100.15 & +0.002 & 37.72 & 43.96\\
Ours Balanced-A & 66.35 & 98.78 & -0.673 & 36.47 & 41.67 & 65.78 & 98.58 & -0.815 & 35.31 & 41.30 & 66.99 & 100.32 & +0.101 & 37.81 & 44.10\\
\bottomrule\end{tabular*}\par
\begin{tablenotes}[flushleft]
\fontsize{7.5}{8.5}\selectfont\raggedright\setlength{\labelsep}{0pt}
\item[] ViCrop inputs both the original image and a crop. The same visual-token cap (256 or 1024) applies to each image.
\item[] N/A: ViCrop has no official InternVL3-8B implementation.
\end{tablenotes}
\end{threeparttable}
\end{table*}

\subsection{\texorpdfstring{Experimental Setup}{Experimental Setup}}
\label{sec:exp-setup}
\noindent\textbf{Data and splitting.}\quad
Our pool contains 206,996 questions from 18 datasets~\mbox{\cite{ai2d,chartqa,docvqa,hrbench,infovqa,mme,mmstar,ocrbench,pinpoint,pope,realworldqa,scienceqa,seedbench,sou,textvqa,vizwiz,vstar,zoombench}} covering local detail, global context, text/document understanding, and mixed perception. To prevent data leakage, we group samples from the same image or document and merge groups sharing the same source across datasets. Using a fixed random seed, we divide the groups into training, validation, and test sets in an approximate 4:2:4 ratio, yielding 83,902, 40,874, and 82,220 samples, respectively.

For each target model, we train one router on the pooled training sets from all 18 datasets and use the pooled validation set for model and configuration selection. We then evaluate the selected configuration on each dataset's test set. We average results equally across the 18 datasets. We repeat router training and validation-set configuration selection with four seeds and report their mean results.

\noindent\textbf{Targets and budgets.}\quad
We evaluate frozen Qwen2.5-VL-7B and -3B~\cite{qwen25vl}, and InternVL3-8B~\cite{internvl3}, with all training and evaluation on RTX 3090 GPUs. To accommodate consumer GPU resources and measure routing gains beyond budget restriction, the original-input and crop paths use a 1,024-token cap; compression uses 256- or 512-token caps. M and A denote two router variants: the selection head uses shared image--question features in M and additionally receives a question-derived task category in A. We report Safe and Balanced configurations, prioritizing accuracy retention and computational savings, respectively.

\noindent\textbf{Comparisons.}\quad
We compare with CARES~\cite{cares}, CropVLM~\cite{cropvlm}, ViCrop~\cite{vicrop}, and ZoomEye~\cite{zoomeye}. CARES is our closest baseline: it shares the goal of reducing computation and visual tokens while preserving performance, and uses a model-agnostic frontend. We cap it at 1,024 visual tokens and retrain its resolution selector, denoting this version CARES-F1024. CropVLM combines an original image and a learned crop. We cap both at 512 visual tokens each, giving a total cap of 1,024, and denote this setting CropVLM-F1024. Its crop constructor is selected on validation data, and its cost is included. The selected constructor uses a 2,048-pixel longest-edge setting for localization. ViCrop and ZoomEye primarily improve fine-detail perception through cropping; ZoomEye is also model-agnostic. We retain their multi-image inputs and repeated calls, evaluating ViCrop at 256 or 1,024 tokens per image and ZoomEye at up to 1,024. All calls count toward end-to-end FLOPs. ViCrop uses the official Qwen2.5-VL relative-attention implementation; no official InternVL3-8B implementation was available.

\subsection{Preliminary Study}
\label{sec:preliminary}
To distinguish method gains from budget effects, we first examine a visual-token cap alone, without adaptive cropping or compression. We evaluate Qwen2.5-VL-7B on 672 questions from four representative datasets---V*~\cite{vstar}, DocVQA~\cite{docvqa}, RealWorldQA~\cite{realworldqa}, and AI2D~\cite{ai2d}. Lowering the cap from 16,384 to 1,024 with other inference settings fixed saves 55.64\% FLOPs and 54.32\% visual tokens, with a 0.63-point mean score loss. Peak allocated GPU memory for target inference falls from 18.81 to 15.75 GiB. This saving with limited average score loss motivates our 1,024-token original-input baseline, against which we measure each method's additional FLOPs and token savings (Appendix~\ref{sec:s-budget-reference}).

\subsection{Overall Performance}
\label{sec:overall}

As shown in Table~\ref{tab:main}, with the original input already capped at 1,024 visual tokens, our method generally reduces end-to-end computation and visual-token consumption while largely retaining task performance across three target models and 18 datasets. Balanced-A saves an average of 36.53\% of end-to-end FLOPs and 42.36\% of visual tokens while retaining 99.23\% of the task score. Safe configurations prioritize accuracy retention: Safe-A saves 19.35--25.32\% of end-to-end FLOPs across the three models, with mean score decreases of 0.19 and less than 0.01 percentage points on Qwen2.5-VL-7B and -3B, respectively, and a gain of 0.85 points on InternVL3-8B. Full per-dataset results are provided in Appendix~\ref{sec:s-per-dataset}. Generalization to held-out datasets is evaluated in Appendix~\ref{sec:s-heldout}.

Compared with CARES-F1024~\cite{cares}, Balanced-A achieves a higher mean task score and greater end-to-end FLOPs and visual-token savings on all three models (Appendix~\ref{sec:s-tradeoff}). Across models, mean task score improves by 1.11 percentage points, and FLOPs savings relative to the original input rise from 32.87\% to 36.53\%.

Per-dataset results show smaller worst-case losses and variation than CARES-F1024 (Appendix~\ref{sec:s-stability}). Across models, CARES-F1024's largest gains are 0.72--2.97 points and its largest losses 5.40--13.83 points. Balanced-A's largest gains are 3.92--9.16 points and losses 3.03--4.32 points; Safe-A's are 2.93--9.65 and 1.30--2.21 points, respectively. Both routing configurations have lower cross-dataset variation than CARES-F1024 on all three models, with Safe configurations limiting the largest losses.

CropVLM-F1024 retains 97.40\%, 97.82\%, and 98.55\% of the task score on 7B, 3B, and InternVL3-8B, but increases end-to-end FLOPs by 35.93\%, 56.23\%, and 49.68\%, respectively. These costs include generating the crop and running the VLM on both the original image and crop.

ViCrop~\cite{vicrop} and ZoomEye~\cite{zoomeye} illustrate a different tradeoff, improving visual perception through additional computation. With 1,024 visual tokens per image, ViCrop improves mean task scores by 0.49--0.97 percentage points on the two Qwen models, but uses approximately $3.03\times$ the end-to-end FLOPs of the original input. ZoomEye improves scores by 2.07--3.23 points across the three models at $24.29$--$53.79\times$ the computation. Although ViCrop's 256-token setting reduces visual tokens in the final answer call, its end-to-end computation remains higher than the original input and its task scores decrease.

\begin{figure}[t]
\centering
\includegraphics[width=\linewidth]{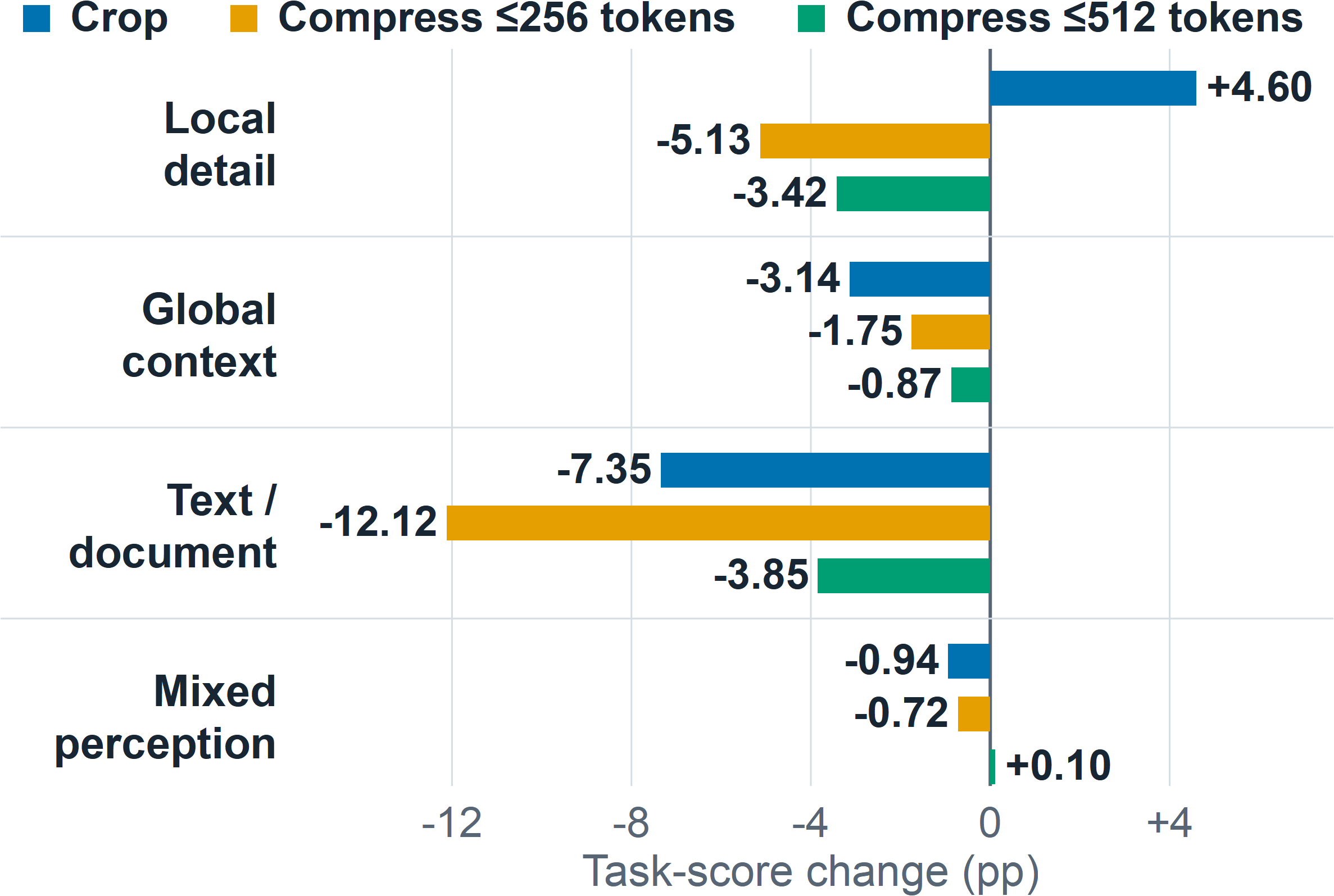}
\caption{Average score changes (pp) across task groups on Qwen2.5-VL-7B.}
\label{fig:complement}
\end{figure}

\subsection{\texorpdfstring{Complementary Cropping and Compression}{Complementary Cropping and Compression}}
\label{sec:crop-or-compress}

Fig.~\ref{fig:complement} compares average score changes relative to the original input for cropping and two whole-image compression budgets across four task groups on Qwen2.5-VL-7B. On local-detail tasks, cropping improves mean score by 4.60 percentage points, whereas $C_{256}$ and $C_{512}$ reduce it by 5.13 and 3.42 points. Global-context tasks show a different pattern: cropping reduces mean score by 3.14 points, compared with 1.75 and 0.87 points for the two compression budgets. These results demonstrate the complementary strengths of cropping and compression.

\begin{table}[t]
\caption{Cropping and compression branch ablation.}
\label{tab:branch-summary}
\centering\fontsize{7.5}{8.5}\selectfont\setlength{\tabcolsep}{2pt}
\begin{tabular}{@{}cccc@{}}\toprule
\multicolumn{4}{c}{\textbf{\boldmath Score change (pp) / FLOPs saving (\%)}}\\
\textbf{\boldmath Target} & \textbf{\boldmath Full router} & \textbf{\boldmath No crop} & \textbf{\boldmath No compression}\\\midrule
7B & -0.535 / 39.09 & -1.994 / 39.81 & +0.867 / 9.01\\
3B & -0.862 / 35.18 & -1.185 / 34.38 & +0.567 / -0.09\\
InternVL3-8B & +0.104 / 37.75 & -0.014 / 37.93 & +0.680 / -3.60\\
\bottomrule\end{tabular}
\end{table}

\begin{table}[t]
\caption{\textbf{Safety-gate ablation.} Cells show score change (pp) / end-to-end FLOPs savings (\%).}
\label{tab:safety-summary}
\centering
\fontsize{7.5}{8.5}\selectfont\setlength{\tabcolsep}{1.5pt}
\begin{tabular}{@{}ccccc@{}}\toprule
\textbf{\boldmath Gate} & \textbf{\boldmath 7B} & \textbf{\boldmath 3B} & \textbf{\boldmath InternVL3-8B} & \textbf{\boldmath \makecell[cc]{Mean\\(3 models)}}\\\midrule
\multicolumn{5}{c}{\textbf{Balanced-A}}\\
None & -1.206 / 42.33 & -1.986 / 43.51 & -1.556 / 42.76 & -1.583 / 42.87\\
Crop & -1.292 / 40.93 & -1.933 / 42.50 & -1.571 / 42.72 & -1.599 / 42.05\\
Compression & -0.587 / 37.87 & -0.868 / 36.33 & +0.116 / 37.84 & -0.446 / 37.35\\
Both & -0.673 / 36.47 & -0.815 / 35.31 & +0.101 / 37.81 & -0.462 / 36.53\\
\midrule
\multicolumn{5}{c}{\textbf{Safe-A}}\\
None & -1.211 / 41.96 & -1.752 / 39.88 & -0.980 / 40.88 & -1.314 / 40.91\\
Crop & -1.281 / 38.29 & -1.561 / 36.42 & -0.931 / 40.62 & -1.258 / 38.45\\
Compression & -0.119 / 28.99 & -0.193 / 22.81 & +0.800 / 25.51 & +0.163 / 25.77\\
Both & -0.189 / 25.32 & -0.002 / 19.35 & +0.849 / 25.26 & +0.219 / 23.31\\
\bottomrule\end{tabular}
\end{table}

\begin{table}[t]
\caption{FLOPs breakdown and savings for Balanced-A.}
\label{tab:compute-breakdown}
\centering
\fontsize{7.5}{8.5}\selectfont\setlength{\tabcolsep}{1.3pt}
\begin{tabular}{@{}ccccccc@{}}\toprule
 & \multicolumn{4}{c}{\textbf{\boldmath Stage shares (\%)}} & \multicolumn{2}{c}{\textbf{\boldmath FLOPs savings (\%)}}\\
\cmidrule(lr){2-5}\cmidrule(lr){6-7}
\textbf{\boldmath Target} & \textbf{\boldmath \makecell[cc]{Shared\\features}} & \textbf{\boldmath \makecell[cc]{Grounding\\DINO}} & \textbf{\boldmath \makecell[cc]{Prediction\\heads}} & \textbf{\boldmath \makecell[cc]{Final\\answer}} & \textbf{\boldmath \makecell[cc]{Final\\answer}} & \textbf{\boldmath End-to-end}\\\midrule
7B & 2.60 & 0.68 & $<0.01$ & 96.71 & 39.60 & 36.47 \\
3B & 3.95 & 0.71 & $<0.01$ & 95.34 & 40.20 & 35.31 \\
InternVL3-8B & 3.04 & 0.01 & $<0.01$ & 96.94 & 40.02 & 37.81 \\
\bottomrule\end{tabular}
\end{table}

\begin{table}[t]
\caption{VisionZip combinations at 70\% token retention on Qwen2.5-VL-7B, averaged over 18 datasets.}
\label{tab:visionzip-composition}
\centering
\fontsize{7.5}{8.5}\selectfont
\renewcommand{\arraystretch}{0.95}
\setlength{\tabcolsep}{2pt}
\begin{threeparttable}
\begin{tabular*}{\linewidth}{@{\extracolsep{\fill}}cccc@{}}\toprule
\textbf{\boldmath Method} & \makecell[c]{\textbf{\boldmath $\Delta$score}\\\textbf{(pp)}} & \makecell[c]{\textbf{FLOPs}\\\textbf{saved (\%)}} & \makecell[c]{\textbf{Tokens}\\\textbf{saved (\%)}}\\\midrule
Original input & +0.00 & 0.00 & 0.00 \\
Ours & -0.55 & 39.09 & 44.27 \\
VisionZip & -0.28 & 20.39 & 30.13 \\
Ours + VisionZip (direct) & -1.38 & 50.29 & 61.13 \\
Ours + VisionZip (adapted) & +0.37 & 29.44 & 42.15 \\
\bottomrule\end{tabular*}
\begin{tablenotes}[flushleft]
\fontsize{7.5}{8.5}\selectfont\raggedright\setlength{\labelsep}{0pt}
\item[] Ours: Balanced-A. Original input and Ours are unpruned. Changes and savings are relative to original input; FLOPs include frontend computation.
\end{tablenotes}
\end{threeparttable}
\end{table}

Text/document tasks are more sensitive to input reduction. Cropping, $C_{256}$, and $C_{512}$ reduce mean score by 7.35, 12.12, and 3.85 points, respectively, with stronger compression causing the largest loss. Thus, both reduction strategy and strength affect task performance.

On mixed-perception tasks, the average changes are small: cropping and $C_{256}$ reduce score by 0.94 and 0.72 points, while $C_{512}$ improves it by 0.10 points. Moderate compression thus largely preserves average performance while reducing input cost.

\subsection{\texorpdfstring{Ablation Studies}{Ablation Studies}}
\label{sec:exp-ablation}

\noindent\textbf{Cropping and compression branches.}\quad
We ablate the cropping and compression branches in Balanced-A (Table~\mbox{\ref{tab:branch-summary}}). Removing cropping lowers mean local-detail task scores by 4.56, 1.22, and 0.43 percentage points on 7B, 3B, and InternVL3-8B, respectively, compared with only 0.12, 0.01, and 0.00 points on global-context tasks. Cropping's benefits are therefore concentrated in local-detail tasks. Removing compression reduces computational savings on all 18 datasets for each model, leaving mean end-to-end FLOPs savings of 9.01\%, -0.09\%, and -3.60\%, respectively. The branches thus make different contributions: cropping improves local-detail performance, while compression provides computational savings across a broader range of tasks. Task-group results are provided in Appendix~\ref{sec:s-whole-branch}.

\noindent\textbf{Safety gate.}\quad
We conduct an ablation study on the safety gate (Table~\mbox{\ref{tab:safety-summary}}). In Balanced-A, enabling both gates reduces the three-model mean score loss from 1.583 to 0.462 points, while FLOPs saving changes from 42.87\% to 36.53\%. The compression gate provides most of the accuracy recovery. Adding the crop gate slightly lowers the mean Balanced-A score but improves Safe-A by 0.057 points on average, including 0.191 points on 3B. Complete ablations are provided in Appendix~\ref{sec:s-safety}.

\subsection{\texorpdfstring{FLOPs, Latency, and GPU Memory}{FLOPs, Latency, and GPU Memory}}
\label{sec:resources}
\noindent\textbf{Computational breakdown.}\quad
Target VLM inference uses 95.34--96.94\% of end-to-end FLOPs (Table~\mbox{\ref{tab:compute-breakdown}}). Shared feature extraction dominates frontend cost at 2.60--3.95\%; the prediction heads together use less than 0.01\%.

Enabling the gate reduces mean end-to-end FLOPs savings by about 6.3 percentage points (Sec.~\ref{sec:exp-ablation}), mainly by running the target VLM on original rather than reduced inputs; frontend computation barely changes. Further savings therefore depend on avoiding unnecessary fallback while preserving accuracy, rather than reducing the already negligible head computation.

\noindent\textbf{Latency.}\quad
On one RTX 3090, we time Qwen2.5-VL-7B on 180 questions (10 per dataset) with three repeats. Mean latency, including preprocessing and the frontend, is 0.765, 0.618, and 0.659 s for F1024, CARES-F1024, and Balanced-A, respectively. Relative to F1024, Balanced-A reduces latency by 13.72\% averaged across datasets (95\% CI: 9.47--17.52\%). Compared with CARES, its faster target-model stage is offset by higher frontend overhead. Details are in Appendix~\ref{sec:s-latency}.

\noindent\textbf{Deployment memory.}\quad
With a visual-token cap of 1,024 and batch size 1, the complete pipeline runs on a single 24GB RTX 3090 for all three targets, with both the router frontend and target VLM loaded on the same GPU. Peak allocated GPU memory is 11.01--19.43 GiB. Measurement settings and per-model results are in Appendix~\ref{sec:s-pipeline-memory}.

\subsection{Combination with Visual Token Pruning}
\label{sec:pruning-combination}

\method\ processes images before they enter the VLM, whereas token pruning operates inside the model. Since these methods act at different stages, they can be applied together. We test this combination using VisionZip~\cite{visionzip} on Qwen2.5-VL-7B at 70\% token retention. Direct combination keeps our router unchanged; adaptation retrains its frontend heads using the pruned VLM's answers.

As shown in Table~\ref{tab:visionzip-composition}, adding our router to VisionZip increases end-to-end FLOPs savings from 20.39\% to 50.29\% and visual-token savings from 30.13\% to 61.13\%, with an additional 1.10-point score loss. With adaptation, the combination improves mean score by 0.65 points over VisionZip alone, while increasing FLOPs savings to 29.44\% and token savings to 42.15\%. Additional results and analysis are provided in Appendix~\ref{sec:s-pruning-combination}.

\section{\texorpdfstring{Limitations and Future Work}{Limitations and Future Work}}
Our main results measure routing gains over a capped baseline. Fully separating savings from the cap and from routing across all models and datasets remains future work, as our cap study covers only one model and four datasets.

The crop safety gate shows potential, but its accuracy gains vary across models and configurations. A more reliable fallback criterion is needed to protect accuracy while preserving beneficial crops.

\section{Conclusion}
We find that a visual-token cap alone substantially reduces computation with a small average performance loss. We therefore measure our method's additional gains over capped original inputs. In this setting, we propose \method, which exploits the complementarity of cropping and compression to reduce computation and visual-token consumption, and uses a safety gate to guide fallback and protect accuracy. We conduct extensive evaluation on 3 VLMs and 18 datasets. With original inputs already capped at 1,024 visual tokens, our main configuration further saves an average of 36.53\% in end-to-end FLOPs and 42.36\% in visual tokens while retaining 99.23\% of the task score. On Qwen2.5-VL-7B, our pipeline also reduces end-to-end latency by 13.72\% on an RTX 3090.

\section*{LLM Usage Statement}
The authors developed the research ideas, experimental plans, manuscript drafts, and figure designs. The authors provided the core relevant references, while OpenAI Codex supplemented the bibliography with additional relevant literature. OpenAI Codex translated and polished the text, produced figures following the authors' designs, and implemented and ran experiments under the authors' direction. The authors iteratively reviewed the manuscript, figures, and results, directed revisions, and take responsibility for the final work. 
\renewcommand{\refname}{\texorpdfstring{References}{References}}
{\small\bibliographystyle{ieeenat_fullname}\bibliography{references}}
\clearpage
\appendix
\setcounter{table}{0}
\setcounter{figure}{0}
\setcounter{equation}{0}
\renewcommand{\thetable}{S\arabic{table}}
\renewcommand{\thefigure}{S\arabic{figure}}
\renewcommand{\theequation}{S\arabic{equation}}
\renewcommand{\theHtable}{S\arabic{table}}
\renewcommand{\theHfigure}{S\arabic{figure}}
\renewcommand{\theHequation}{S\arabic{equation}}
\section{Complete Results}
\label{sec:s-results}
Scores use each dataset's metric on a 0--100 scale. Score changes are percentage points; FLOPs and visual-token savings are percentages relative to the same target's original-input baseline. Ours reports means over four training seeds. In metric columns, D denotes equal averaging over datasets and M denotes equal averaging over task groups; definitions and sample counts are in Sec.~\ref{sec:s-protocol}.

Mechanism-balanced results first average datasets within each group, then average four groups equally: local detail (V*, Pinpoint-Bench, HR-Bench 8K, ZoomBench, SOU); global context (ChartQA, AI2D, MME, RealWorldQA, ScienceQA); text/document (TextVQA, DocVQA, OCRBench, InfoVQA); and mixed perception (VizWiz, MMStar, SEED-Bench Image, POPE). The grouping was recorded after the initial core results and serves as retrospective sensitivity analysis.

\subsection{Aggregate results}
Table~\ref{tab:operating-points} reports all 12 model/configuration combinations and their three-model means. Validation counts indicate how many runs satisfy the corresponding quality limits; all runs contribute to the reported results. Fixed-input, mechanism-balanced method, and CARES results appear in Tables~\ref{tab:aggregate-results}--\ref{tab:cares-comparison}.
\begin{table*}[!t]
\caption{Complete routing results and validation feasibility across four seeds. D / M denotes dataset-macro / mechanism-balanced macro. Counts are out of four per target and twelve for the three-target mean.}\label{tab:operating-points}
\centering\fontsize{7}{8}\selectfont\setlength{\tabcolsep}{3pt}
\begin{tabular}{cccccccc}\toprule
\textbf{\boldmath Target} & \textbf{\boldmath Setting} & \textbf{\boldmath \makecell[cc]{Validation\\met}} & \textbf{\boldmath \makecell[cc]{Score\\D / M}} & \textbf{\boldmath \makecell[cc]{Retention (\%)\\D / M}} & \textbf{\boldmath \makecell[cc]{$\Delta$pp\\D / M}} & \textbf{\boldmath \makecell[cc]{FLOPs saving (\%)\\D / M}} & \textbf{\boldmath \makecell[cc]{Token saving (\%)\\D / M}}\\\midrule
7B & Safe-M & 4/4 & 66.79 / 67.39 & 99.44 / 99.48 & -0.234 / -0.225 & 25.35 / 24.92 & 30.42 / 30.02\\
7B & Safe-A & 4/4 & 66.84 / 67.43 & 99.48 / 99.52 & -0.189 / -0.184 & 25.32 / 24.87 & 30.52 / 30.10\\
7B & Balanced-M & 4/4 & 66.11 / 66.71 & 98.40 / 98.47 & -0.911 / -0.903 & 37.39 / 37.04 & 42.34 / 41.99\\
7B & Balanced-A & 4/4 & 66.35 / 66.92 & 98.78 / 98.81 & -0.673 / -0.692 & 36.47 / 36.25 & 41.67 / 41.46\\
3B & Safe-M & 4/4 & 66.70 / 67.35 & 100.19 / 100.11 & +0.099 / +0.044 & 21.52 / 20.80 & 28.92 / 28.29\\
3B & Safe-A & 4/4 & 66.59 / 67.28 & 99.93 / 99.89 & -0.002 / -0.034 & 19.35 / 18.69 & 26.03 / 25.46\\
3B & Balanced-M & 4/4 & 65.71 / 66.37 & 98.42 / 98.39 & -0.887 / -0.941 & 35.21 / 34.45 & 41.13 / 40.42\\
3B & Balanced-A & 4/4 & 65.78 / 66.43 & 98.58 / 98.53 & -0.815 / -0.878 & 35.31 / 34.55 & 41.30 / 40.59\\
InternVL3-8B & Safe-M & 2/4 & 67.75 / 68.24 & 101.44 / 101.51 & +0.864 / +0.895 & 25.38 / 25.14 & 30.84 / 30.51\\
InternVL3-8B & Safe-A & 3/4 & 67.74 / 68.22 & 101.41 / 101.48 & +0.849 / +0.882 & 25.26 / 25.02 & 30.73 / 30.40\\
InternVL3-8B & Balanced-M & 4/4 & 66.89 / 67.37 & 100.15 / 100.23 & +0.002 / +0.028 & 37.72 / 37.32 & 43.96 / 43.46\\
InternVL3-8B & Balanced-A & 4/4 & 66.99 / 67.46 & 100.32 / 100.38 & +0.101 / +0.117 & 37.81 / 37.40 & 44.10 / 43.59\\
Three-target mean & Safe-M & 10/12 & 67.08 / 67.66 & 100.36 / 100.37 & +0.243 / +0.238 & 24.08 / 23.62 & 30.06 / 29.61\\
Three-target mean & Safe-A & 11/12 & 67.06 / 67.64 & 100.27 / 100.30 & +0.219 / +0.221 & 23.31 / 22.86 & 29.09 / 28.65\\
Three-target mean & Balanced-M & 12/12 & 66.24 / 66.82 & 98.99 / 99.03 & -0.599 / -0.605 & 36.77 / 36.27 & 42.48 / 41.96\\
Three-target mean & Balanced-A & 12/12 & 66.37 / 66.94 & 99.23 / 99.24 & -0.462 / -0.484 & 36.53 / 36.07 & 42.36 / 41.88\\
\bottomrule\end{tabular}
\end{table*}

\begin{table*}[!t]
\caption{Fixed-input score and end-to-end FLOPs saving (\%) across 18 test datasets, with three-target mean changes relative to each target's $F_{1024}$.}
\label{tab:aggregate-results}
\centering
\setlength{\tabcolsep}{3pt}
\scriptsize
\begin{tabular}{ccccccccc}
\toprule
 & \multicolumn{2}{c}{\textbf{\boldmath Qwen2.5-VL-7B}} & \multicolumn{2}{c}{\textbf{\boldmath Qwen2.5-VL-3B}} & \multicolumn{2}{c}{\textbf{\boldmath InternVL3-8B}} & \multicolumn{2}{c}{\textbf{\boldmath Target mean}} \\
\midrule
Method & Score & Saving$\uparrow$ & Score & Saving$\uparrow$ & Score & Saving$\uparrow$ & $\Delta$pp & Saving$\uparrow$ \\
$F_{1024}$ & 67.03 & 0.00 & 66.60 & 0.00 & 66.89 & 0.00 & +0.000 & 0.00 \\
$R_{1024}$ & 65.59 & 17.46 & 64.63 & 14.29 & 65.37 & 2.01 & -1.639 & 11.25 \\
$C_{256}$ & 62.26 & 63.10 & 60.77 & 64.50 & 63.79 & 55.56 & -4.564 & 61.05 \\
$C_{512}$ & 65.00 & 38.81 & 64.60 & 39.80 & 65.24 & 36.64 & -1.891 & 38.42 \\
\bottomrule\end{tabular}

\end{table*}

\begin{table*}[!t]
\caption{Mechanism-balanced score retention and savings across 18 datasets (\%). N/A denotes unevaluated ViCrop on InternVL3-8B.}
\label{tab:method-comparison-mechanism}
\centering\scriptsize\setlength{\tabcolsep}{3pt}
\begin{threeparttable}
\begin{tabular}{cccccccccc}\toprule
 & \multicolumn{3}{c}{\textbf{\boldmath Qwen2.5-VL-7B}} & \multicolumn{3}{c}{\textbf{\boldmath Qwen2.5-VL-3B}} & \multicolumn{3}{c}{\textbf{\boldmath InternVL3-8B}}\\
\textbf{\boldmath Method} & \textbf{\boldmath Retention} & \textbf{\boldmath \makecell[cc]{FLOPs\\saved}} & \textbf{\boldmath \makecell[cc]{Tokens\\saved}} & \textbf{\boldmath Retention} & \textbf{\boldmath \makecell[cc]{FLOPs\\saved}} & \textbf{\boldmath \makecell[cc]{Tokens\\saved}} & \textbf{\boldmath Retention} & \textbf{\boldmath \makecell[cc]{FLOPs\\saved}} & \textbf{\boldmath \makecell[cc]{Tokens\\saved}}\\\midrule
\multicolumn{10}{c}{\textbf{All 18 datasets}}\\
CARES-F1024~\cite{cares} & 97.76 & 35.75 & 39.39 & 96.51 & 32.93 & 37.01 & 98.75 & 29.82 & 34.85 \\
CropVLM-F1024~\cite{cropvlm} & 97.41 & -36.17 & -8.78 & 97.68 & -56.58 & -8.78 & 98.39 & -49.65 & -21.62\\
ViCrop 256~\cite{vicrop} & 94.21 & -21.22 & 56.06 & 94.25 & -18.08 & 56.04 & \multicolumn{3}{c}{N/A}\\
ViCrop 1024 & 101.97 & -203.13 & -5.43 & 101.23 & -203.30 & -5.24 & \multicolumn{3}{c}{N/A}\\
ZoomEye official final-cap~\cite{zoomeye} & 105.01 & -2182.36 & 9.58 & 105.98 & -4436.00 & 16.45 & 107.70 & -5036.86 & 1.56 \\
Ours Safe-A & 99.52 & 24.87 & 30.10 & 99.89 & 18.69 & 25.46 & 101.48 & 25.02 & 30.40\\
Ours Balanced-A & 98.81 & 36.25 & 41.46 & 98.53 & 34.55 & 40.59 & 100.38 & 37.40 & 43.59\\
\bottomrule\end{tabular}
\begin{tablenotes}[flushleft]
\fontsize{7.5}{8.5}\selectfont\raggedright\setlength{\labelsep}{0pt}
\item[] ViCrop inputs both the original image and a crop. The same visual-token cap (256 or 1024) applies to each image.
\end{tablenotes}
\end{threeparttable}
\end{table*}

\begin{table*}[!t]
\caption{CARES and original-input baselines across 18 datasets. D / M denotes dataset-macro / mechanism-balanced macro; savings are relative to each target's $F_{1024}$.}\label{tab:cares-comparison}
\centering\scriptsize\setlength{\tabcolsep}{3pt}
\begin{tabular}{cccccc}\toprule
\textbf{\boldmath Target} & \textbf{\boldmath Method} & \textbf{\boldmath \makecell[cc]{Score\\D / M}} & \textbf{\boldmath \makecell[cc]{$\Delta$pp\\D / M}} & \textbf{\boldmath \makecell[cc]{FLOPs saving (\%)\\D / M}} & \textbf{\boldmath \makecell[cc]{Token saving (\%)\\D / M}}\\\midrule
7B & $F_{1024}$ & 67.03 / 67.61 & +0.000 / +0.000 & 0.00 / 0.00 & 0.00 / 0.00 \\
7B & CARES-F1024~\cite{cares} & 65.65 / 66.22 & -1.373 / -1.391 & 35.69 / 35.75 & 39.36 / 39.39 \\
3B & $F_{1024}$ & 66.60 / 67.31 & +0.000 / +0.000 & 0.00 / 0.00 & 0.00 / 0.00 \\
3B & CARES-F1024~\cite{cares} & 64.33 / 65.03 & -2.265 / -2.277 & 33.03 / 32.93 & 37.12 / 37.01 \\
InternVL3-8B & $F_{1024}$ & 66.89 / 67.34 & +0.000 / +0.000 & 0.00 / 0.00 & 0.00 / 0.00 \\
InternVL3-8B & CARES-F1024~\cite{cares} & 65.81 / 66.15 & -1.077 / -1.189 & 29.89 / 29.82 & 34.98 / 34.85 \\
\bottomrule\end{tabular}
\end{table*}
\subsection{Per-dataset results}
\label{sec:s-per-dataset}
Table~\ref{tab:detailed-results} gives the original input and all four routing configurations on each of the 18 datasets. Tables~\ref{tab:comparators-7b}--\ref{tab:comparators-internvl3-8b} give fixed cropping/compression and external comparisons for the three targets. Negative savings indicate higher computation than the original-input baseline.
\begin{table*}[!t]
\caption{Per-dataset scores and end-to-end FLOPs savings (\%) for all four configurations, relative to each target's $F_{1024}$. $\dagger$ marks InternVL3-8B Safe-M, feasible in 2/4 seeds.}
\label{tab:detailed-results}
\centering
\fontsize{7}{8}\selectfont
\setlength{\tabcolsep}{1.3pt}
\renewcommand{\arraystretch}{1.08}
\begin{tabular*}{\textwidth}{@{\extracolsep{\fill}}ccccccccccccccccccc@{}}\toprule
\multicolumn{19}{c}{\textbf{(a) Datasets 1--9}}\\
 & \multicolumn{2}{c}{\textbf{\boldmath AI2D}} & \multicolumn{2}{c}{\textbf{\boldmath ChartQA}} & \multicolumn{2}{c}{\textbf{\boldmath DocVQA}} & \multicolumn{2}{c}{\textbf{\boldmath \makecell[cc]{HR-Bench\\8K}}} & \multicolumn{2}{c}{\textbf{\boldmath InfoVQA}} & \multicolumn{2}{c}{\textbf{\boldmath MME}} & \multicolumn{2}{c}{\textbf{\boldmath MMStar}} & \multicolumn{2}{c}{\textbf{\boldmath OCRBench}} & \multicolumn{2}{c}{\textbf{\boldmath \makecell[cc]{Pinpoint-\\Bench}}}\\
\cmidrule(lr){2-3}\cmidrule(lr){4-5}\cmidrule(lr){6-7}\cmidrule(lr){8-9}\cmidrule(lr){10-11}\cmidrule(lr){12-13}\cmidrule(lr){14-15}\cmidrule(lr){16-17}\cmidrule(lr){18-19}
\textbf{\boldmath Method} & \textbf{\boldmath Score} & \textbf{\boldmath Saved} & \textbf{\boldmath Score} & \textbf{\boldmath Saved} & \textbf{\boldmath Score} & \textbf{\boldmath Saved} & \textbf{\boldmath Score} & \textbf{\boldmath Saved} & \textbf{\boldmath Score} & \textbf{\boldmath Saved} & \textbf{\boldmath Score} & \textbf{\boldmath Saved} & \textbf{\boldmath Score} & \textbf{\boldmath Saved} & \textbf{\boldmath Score} & \textbf{\boldmath Saved} & \textbf{\boldmath Score} & \textbf{\boldmath Saved}\\\midrule
\multicolumn{19}{c}{\textbf{Qwen2.5-VL-7B}}\\
$F_{1024}$ & 81.25 & 0.00 & 76.51 & 0.00 & 92.83 & 0.00 & 53.72 & 0.00 & 71.17 & 0.00 & 84.54 & 0.00 & 60.45 & 0.00 & 73.51 & 0.00 & 27.08 & 0.00\\
Safe-M & 81.89 & 34.28 & 75.83 & 19.52 & 91.69 & 22.56 & 53.46 & 23.52 & 71.32 & -0.26 & 83.97 & 41.53 & 61.37 & 14.98 & 74.32 & 18.64 & 25.39 & 35.73\\
Safe-A & 81.90 & 33.11 & 75.81 & 18.48 & 91.69 & 22.54 & 53.66 & 22.84 & 71.32 & -0.26 & 83.94 & 41.66 & 61.32 & 14.14 & 74.44 & 18.44 & 24.87 & 35.51\\
Balanced-M & 81.67 & 38.32 & 74.83 & 30.74 & 89.41 & 41.77 & 50.40 & 44.30 & 70.75 & 6.88 & 83.92 & 48.24 & 61.59 & 20.49 & 74.69 & 35.59 & 24.48 & 43.29\\
Balanced-A & 81.70 & 34.56 & 75.29 & 24.99 & 89.37 & 45.42 & 51.66 & 41.25 & 70.10 & 12.70 & 84.07 & 44.23 & 61.50 & 19.95 & 74.50 & 35.59 & 24.22 & 42.86\\
\midrule
\multicolumn{19}{c}{\textbf{Qwen2.5-VL-3B}}\\
$F_{1024}$ & 79.27 & 0.00 & 84.18 & 0.00 & 95.42 & 0.00 & 48.67 & 0.00 & 68.12 & 0.00 & 84.54 & 0.00 & 55.75 & 0.00 & 78.71 & 0.00 & 23.96 & 0.00\\
Safe-M & 79.62 & 26.54 & 83.39 & 12.61 & 95.05 & 4.39 & 52.79 & 29.44 & 67.69 & -0.05 & 84.51 & 40.96 & 55.79 & 6.67 & 78.59 & 8.00 & 22.79 & 30.11\\
Safe-A & 79.51 & 25.94 & 83.64 & 8.06 & 95.10 & 3.61 & 51.60 & 18.74 & 67.74 & -0.38 & 84.64 & 33.16 & 55.75 & 11.71 & 79.02 & 9.30 & 22.66 & 29.76\\
Balanced-M & 79.54 & 34.51 & 81.71 & 31.64 & 93.25 & 19.14 & 52.19 & 45.96 & 65.83 & 12.42 & 83.89 & 55.08 & 55.53 & 17.04 & 77.72 & 25.48 & 21.88 & 46.47\\
Balanced-A & 79.54 & 34.32 & 81.72 & 31.76 & 93.25 & 19.16 & 52.59 & 45.23 & 65.81 & 12.41 & 83.87 & 54.99 & 55.36 & 16.87 & 77.72 & 25.58 & 22.14 & 45.81\\
\midrule
\multicolumn{19}{c}{\textbf{InternVL3-8B}}\\
$F_{1024}$ & 81.47 & 0.00 & 73.60 & 0.00 & 91.83 & 0.00 & 52.66 & 0.00 & 61.95 & 0.00 & 87.42 & 0.00 & 64.11 & 0.00 & 64.85 & 0.00 & 26.56 & 0.00\\
Safe-M$^{\dagger}$ & 81.78 & 37.87 & 74.04 & 37.07 & 90.80 & 4.46 & 50.66 & 19.23 & 61.33 & 2.12 & 88.92 & 45.20 & 65.37 & 14.11 & 74.50 & 29.58 & 26.82 & 12.84\\
Safe-A & 81.78 & 37.87 & 74.04 & 37.06 & 90.80 & 4.32 & 50.60 & 19.44 & 61.31 & 2.07 & 88.92 & 45.20 & 65.37 & 14.09 & 74.50 & 29.40 & 26.95 & 12.19\\
Balanced-M & 81.85 & 40.44 & 72.86 & 63.52 & 88.71 & 18.70 & 47.74 & 33.46 & 58.66 & 15.43 & 89.15 & 48.01 & 65.72 & 16.33 & 74.01 & 43.02 & 27.86 & 38.31\\
Balanced-A & 81.85 & 40.44 & 72.86 & 63.52 & 88.71 & 18.69 & 48.34 & 32.87 & 58.66 & 15.43 & 89.15 & 48.01 & 65.72 & 16.33 & 74.01 & 43.02 & 27.86 & 38.31\\
\bottomrule\end{tabular*}
\end{table*}
\begin{table*}[!t]
\centering
{\small Table~\ref{tab:detailed-results} (continued).}\par\vspace{5pt}
\fontsize{7}{8}\selectfont
\setlength{\tabcolsep}{1.3pt}
\renewcommand{\arraystretch}{1.08}
\begin{tabular*}{\textwidth}{@{\extracolsep{\fill}}ccccccccccccccccccc@{}}\toprule
\multicolumn{19}{c}{\textbf{(b) Datasets 10--18}}\\
 & \multicolumn{2}{c}{\textbf{\boldmath POPE}} & \multicolumn{2}{c}{\textbf{\boldmath \makecell[cc]{RealWorld\\QA}}} & \multicolumn{2}{c}{\textbf{\boldmath ScienceQA}} & \multicolumn{2}{c}{\textbf{\boldmath \makecell[cc]{SEED-Bench\\Image}}} & \multicolumn{2}{c}{\textbf{\boldmath SOU}} & \multicolumn{2}{c}{\textbf{\boldmath TextVQA}} & \multicolumn{2}{c}{\textbf{\boldmath VizWiz}} & \multicolumn{2}{c}{\textbf{\boldmath V$^*$}} & \multicolumn{2}{c}{\textbf{\boldmath ZoomBench}}\\
\cmidrule(lr){2-3}\cmidrule(lr){4-5}\cmidrule(lr){6-7}\cmidrule(lr){8-9}\cmidrule(lr){10-11}\cmidrule(lr){12-13}\cmidrule(lr){14-15}\cmidrule(lr){16-17}\cmidrule(lr){18-19}
\textbf{\boldmath Method} & \textbf{\boldmath Score} & \textbf{\boldmath Saved} & \textbf{\boldmath Score} & \textbf{\boldmath Saved} & \textbf{\boldmath Score} & \textbf{\boldmath Saved} & \textbf{\boldmath Score} & \textbf{\boldmath Saved} & \textbf{\boldmath Score} & \textbf{\boldmath Saved} & \textbf{\boldmath Score} & \textbf{\boldmath Saved} & \textbf{\boldmath Score} & \textbf{\boldmath Saved} & \textbf{\boldmath Score} & \textbf{\boldmath Saved} & \textbf{\boldmath Score} & \textbf{\boldmath Saved}\\\midrule
\multicolumn{19}{c}{\textbf{Qwen2.5-VL-7B}}\\
$F_{1024}$ & 85.54 & 0.00 & 62.75 & 0.00 & 82.89 & 0.00 & 76.53 & 0.00 & 53.31 & 0.00 & 80.53 & 0.00 & 42.50 & 0.00 & 66.67 & 0.00 & 34.68 & 0.00\\
Safe-M & 85.85 & 20.67 & 61.19 & 19.19 & 81.98 & 34.01 & 75.77 & 30.63 & 53.33 & 32.04 & 79.13 & 17.91 & 42.50 & 43.39 & 69.44 & 25.13 & 33.82 & 22.84\\
Safe-A & 85.77 & 20.62 & 61.19 & 20.94 & 82.11 & 34.18 & 75.79 & 30.43 & 53.45 & 31.88 & 79.17 & 17.51 & 42.49 & 43.30 & 69.79 & 27.50 & 34.32 & 23.00\\
Balanced-M & 85.46 & 26.19 & 60.78 & 47.29 & 82.33 & 35.37 & 75.06 & 52.57 & 53.07 & 37.30 & 76.97 & 41.30 & 42.49 & 46.65 & 67.71 & 37.43 & 34.47 & 39.28\\
Balanced-A & 85.72 & 25.96 & 60.21 & 44.20 & 82.20 & 34.60 & 75.28 & 48.53 & 52.87 & 37.26 & 77.19 & 39.23 & 42.50 & 46.46 & 70.83 & 38.11 & 35.12 & 40.65\\
\midrule
\multicolumn{19}{c}{\textbf{Qwen2.5-VL-3B}}\\
$F_{1024}$ & 87.56 & 0.00 & 57.52 & 0.00 & 74.72 & 0.00 & 74.83 & 0.00 & 52.65 & 0.00 & 82.05 & 0.00 & 47.38 & 0.00 & 65.28 & 0.00 & 38.15 & 0.00\\
Safe-M & 87.42 & 8.92 & 57.35 & 25.70 & 73.90 & 24.02 & 73.72 & 36.84 & 52.06 & 35.66 & 81.20 & 7.85 & 46.74 & 42.10 & 68.75 & 36.08 & 39.16 & 11.51\\
Safe-A & 87.41 & 4.60 & 56.29 & 28.02 & 74.18 & 21.82 & 74.17 & 25.17 & 51.93 & 34.16 & 81.29 & 7.60 & 46.77 & 40.95 & 68.06 & 33.17 & 38.95 & 12.83\\
Balanced-M & 86.73 & 19.30 & 57.19 & 45.49 & 73.96 & 30.96 & 73.40 & 51.00 & 51.08 & 41.46 & 79.25 & 30.68 & 46.71 & 46.12 & 67.71 & 39.94 & 35.19 & 41.04\\
Balanced-A & 86.83 & 19.86 & 57.35 & 44.66 & 73.99 & 31.07 & 73.38 & 51.00 & 51.08 & 41.46 & 79.22 & 30.32 & 46.70 & 46.03 & 68.40 & 42.62 & 35.12 & 42.52\\
\midrule
\multicolumn{19}{c}{\textbf{InternVL3-8B}}\\
$F_{1024}$ & 89.39 & 0.00 & 63.73 & 0.00 & 90.31 & 0.00 & 74.90 & 0.00 & 50.04 & 0.00 & 87.83 & 0.00 & 36.47 & 0.00 & 65.28 & 0.00 & 41.62 & 0.00\\
Safe-M$^{\dagger}$ & 89.19 & 15.44 & 63.89 & 13.83 & 92.26 & 38.90 & 75.38 & 48.18 & 50.21 & 21.45 & 86.56 & 37.27 & 37.55 & 32.72 & 67.01 & 19.82 & 43.28 & 26.71\\
Safe-A & 89.19 & 15.44 & 63.64 & 13.55 & 92.26 & 38.90 & 75.39 & 48.23 & 50.21 & 21.45 & 86.62 & 36.77 & 37.56 & 32.55 & 67.36 & 16.52 & 42.77 & 29.56\\
Balanced-M & 89.19 & 20.39 & 63.48 & 26.30 & 92.33 & 40.20 & 75.50 & 59.96 & 49.98 & 31.15 & 83.96 & 58.51 & 37.64 & 37.32 & 64.93 & 45.20 & 40.46 & 42.74\\
Balanced-A & 89.19 & 20.39 & 63.48 & 26.30 & 92.33 & 40.20 & 75.50 & 59.96 & 49.98 & 31.15 & 83.96 & 58.51 & 37.64 & 37.32 & 65.97 & 48.12 & 40.61 & 42.01\\
\bottomrule\end{tabular*}\par

\end{table*}

\begin{table*}[!t]
\caption{Qwen2.5-VL-7B: Per-dataset scores (0--100) and end-to-end FLOPs savings (\%) relative to $F_{1024}$. $R$ denotes crop and $C$ whole-image compression; subscripts are visual-token caps. ZoomEye official final-cap and matched-F fallback are distinct configurations.}
\label{tab:comparators-7b}
\centering\footnotesize\setlength{\tabcolsep}{3pt}
\begin{tabular}{cccccccccccccc}\toprule
 & & \multicolumn{2}{c}{\textbf{\boldmath $R_{1024}$}} & \multicolumn{2}{c}{\textbf{\boldmath $C_{256}$}} & \multicolumn{2}{c}{\textbf{\boldmath $C_{512}$}} & \multicolumn{2}{c}{\textbf{\boldmath CARES}} & \multicolumn{2}{c}{\textbf{\boldmath ZoomEye final-cap}} & \multicolumn{2}{c}{\textbf{\boldmath ZoomEye fallback}} \\
\textbf{\boldmath Dataset} & \textbf{\boldmath $N$} & \textbf{\boldmath Score} & \textbf{\boldmath Saved} & \textbf{\boldmath Score} & \textbf{\boldmath Saved} & \textbf{\boldmath Score} & \textbf{\boldmath Saved} & \textbf{\boldmath Score} & \textbf{\boldmath Saved} & \textbf{\boldmath Score} & \textbf{\boldmath Saved} & \textbf{\boldmath Score} & \textbf{\boldmath Saved} \\\midrule
AI2D & 1365 & 77.58 & 15.54 & 81.61 & 51.00 & 81.61 & 27.08 & 81.90 & 30.21 & 81.17 & -767.30 & 81.17 & -774.10\\
ChartQA & 13038 & 69.81 & 11.33 & 71.84 & 55.92 & 75.93 & 21.28 & 75.73 & 29.98 & 76.50 & -361.47 & 76.47 & -362.36\\
DocVQA & 17484 & 86.56 & 1.08 & 76.69 & 73.52 & 89.11 & 50.32 & 87.61 & 58.65 & 91.89 & -982.04 & 91.79 & -981.99\\
HR-Bench 8K & 376 & 61.17 & 4.14 & 46.01 & 73.18 & 51.06 & 48.65 & 50.27 & 45.92 & 61.97 & -11747.76 & 61.97 & -11747.49\\
InfoVQA & 1098 & 54.16 & 17.86 & 48.81 & 72.44 & 63.83 & 47.06 & 65.77 & 39.80 & 71.20 & -1218.11 & 71.20 & -1218.02\\
MME & 970 & 83.20 & 8.29 & 83.40 & 63.09 & 84.43 & 36.82 & 83.92 & 49.65 & 84.33 & -853.10 & 84.02 & -855.77\\
MMStar & 574 & 57.84 & 20.00 & 60.63 & 34.48 & 61.67 & 22.35 & 62.54 & 14.17 & 60.63 & -234.43 & 61.15 & -251.74\\
OCRBench & 404 & 73.76 & 24.04 & 72.77 & 63.79 & 73.51 & 39.64 & 75.50 & 40.40 & 75.74 & -567.33 & 74.50 & -585.60\\
Pinpoint & 192 & 28.12 & 18.55 & 21.35 & 73.84 & 21.88 & 50.96 & 22.92 & 24.69 & 40.62 & -6925.87 & 40.62 & -6925.87\\
POPE & 3618 & 85.96 & 28.12 & 84.58 & 30.87 & 85.66 & 1.96 & 85.43 & 8.64 & 85.68 & -519.21 & 85.57 & -520.37\\
RealWorldQA & 306 & 62.42 & 24.04 & 59.80 & 72.91 & 59.48 & 45.91 & 59.48 & 44.37 & 61.76 & -1213.57 & 61.76 & -1213.50\\
ScienceQA & 795 & 79.25 & 29.88 & 82.52 & 44.93 & 82.14 & 23.26 & 81.76 & 26.33 & 82.64 & -253.95 & 82.89 & -281.27\\
SEED-Bench Image & 5602 & 76.03 & 12.87 & 74.15 & 71.16 & 75.65 & 46.02 & 75.74 & 49.41 & 76.10 & -1745.16 & 76.10 & -1744.97\\
SOU & 7290 & 51.44 & 5.59 & 51.67 & 63.06 & 52.10 & 41.94 & 53.73 & 29.79 & 53.90 & -1624.31 & 53.16 & -1626.85\\
TextVQA & 15637 & 74.16 & 30.05 & 71.27 & 72.46 & 76.20 & 46.96 & 75.84 & 52.51 & 79.50 & -962.22 & 79.46 & -962.08\\
VizWiz & 13053 & 41.45 & 7.15 & 42.80 & 72.64 & 42.47 & 47.80 & 42.21 & 26.28 & 42.53 & -679.26 & 42.36 & -679.73\\
V$^*$ & 72 & 76.39 & 44.52 & 56.94 & 73.50 & 58.33 & 50.62 & 65.28 & 39.94 & 75.00 & -7241.04 & 75.00 & -7241.04\\
ZoomBench & 346 & 41.33 & 11.24 & 33.82 & 73.01 & 34.97 & 49.99 & 36.13 & 31.74 & 42.49 & -4024.11 & 42.49 & -4024.09\\
\bottomrule\end{tabular}
\end{table*}

\begin{table*}[!t]
\caption{Qwen2.5-VL-3B: Per-dataset scores (0--100) and end-to-end FLOPs savings (\%) relative to $F_{1024}$. $R$ denotes crop and $C$ whole-image compression; subscripts are visual-token caps. ZoomEye official final-cap and matched-F fallback are distinct configurations.}
\label{tab:comparators-3b}
\centering\footnotesize\setlength{\tabcolsep}{3pt}
\begin{tabular}{cccccccccccccc}\toprule
 & & \multicolumn{2}{c}{\textbf{\boldmath $R_{1024}$}} & \multicolumn{2}{c}{\textbf{\boldmath $C_{256}$}} & \multicolumn{2}{c}{\textbf{\boldmath $C_{512}$}} & \multicolumn{2}{c}{\textbf{\boldmath CARES}} & \multicolumn{2}{c}{\textbf{\boldmath ZoomEye final-cap}} & \multicolumn{2}{c}{\textbf{\boldmath ZoomEye fallback}} \\
\textbf{\boldmath Dataset} & \textbf{\boldmath $N$} & \textbf{\boldmath Score} & \textbf{\boldmath Saved} & \textbf{\boldmath Score} & \textbf{\boldmath Saved} & \textbf{\boldmath Score} & \textbf{\boldmath Saved} & \textbf{\boldmath Score} & \textbf{\boldmath Saved} & \textbf{\boldmath Score} & \textbf{\boldmath Saved} & \textbf{\boldmath Score} & \textbf{\boldmath Saved} \\\midrule
AI2D & 1365 & 76.04 & 10.97 & 79.27 & 52.61 & 79.63 & 28.03 & 79.34 & 28.28 & 79.19 & -1974.90 & 79.05 & -1975.35\\
ChartQA & 13038 & 76.02 & 7.70 & 76.56 & 57.47 & 83.08 & 21.97 & 81.63 & 37.46 & 82.31 & -2018.76 & 82.30 & -2018.83\\
DocVQA & 17484 & 89.18 & -1.17 & 70.80 & 74.52 & 88.17 & 51.20 & 89.48 & 50.97 & 90.59 & -4603.96 & 90.58 & -4603.96\\
HR-Bench 8K & 376 & 54.52 & 1.62 & 44.41 & 74.51 & 52.13 & 49.75 & 47.87 & 44.97 & 64.10 & -14299.69 & 64.10 & -14299.69\\
InfoVQA & 1098 & 52.59 & 16.16 & 42.46 & 73.46 & 57.58 & 47.93 & 63.74 & 27.06 & 65.81 & -4666.85 & 65.81 & -4666.85\\
MME & 970 & 83.51 & 4.57 & 83.61 & 64.36 & 84.64 & 37.69 & 85.26 & 44.99 & 84.02 & -3766.97 & 84.02 & -3766.97\\
MMStar & 574 & 53.83 & 14.29 & 56.10 & 36.64 & 55.40 & 23.80 & 55.57 & 12.24 & 56.10 & -2423.70 & 56.27 & -2424.29\\
OCRBench & 404 & 75.99 & 20.66 & 73.02 & 65.63 & 76.98 & 40.83 & 76.73 & 38.96 & 77.97 & -2035.65 & 77.97 & -2035.65\\
Pinpoint & 192 & 27.08 & 16.26 & 20.31 & 74.80 & 21.88 & 51.82 & 22.92 & 23.40 & 40.62 & -9917.92 & 40.62 & -9917.92\\
POPE & 3618 & 86.68 & 21.81 & 86.32 & 31.62 & 87.45 & 2.01 & 87.04 & 3.15 & 87.78 & -1474.30 & 87.78 & -1474.30\\
RealWorldQA & 306 & 59.15 & 21.67 & 56.21 & 74.10 & 58.17 & 46.87 & 54.90 & 40.03 & 58.82 & -3150.53 & 58.82 & -3150.53\\
ScienceQA & 795 & 70.82 & 24.49 & 73.84 & 46.82 & 74.34 & 24.23 & 74.34 & 22.51 & 74.21 & -1076.99 & 74.34 & -1077.93\\
SEED-Bench Image & 5602 & 73.46 & 10.54 & 72.42 & 72.56 & 73.60 & 47.16 & 73.37 & 46.81 & 73.63 & -2851.32 & 73.63 & -2851.32\\
SOU & 7290 & 51.88 & 3.46 & 50.29 & 65.94 & 51.07 & 44.02 & 51.21 & 30.15 & 52.41 & -11174.62 & 52.41 & -11174.62\\
TextVQA & 15637 & 76.35 & 27.98 & 71.56 & 73.44 & 77.88 & 47.79 & 77.94 & 49.27 & 79.84 & -2060.59 & 79.84 & -2060.58\\
VizWiz & 13053 & 45.18 & 4.72 & 45.09 & 73.65 & 46.71 & 48.68 & 46.92 & 27.42 & 44.45 & -3300.36 & 44.47 & -3300.41\\
V$^*$ & 72 & 69.44 & 42.56 & 55.56 & 74.61 & 59.72 & 51.59 & 54.17 & 40.91 & 80.56 & -7177.64 & 80.56 & -7177.64\\
ZoomBench & 346 & 41.62 & 8.93 & 36.13 & 74.27 & 34.39 & 51.05 & 35.55 & 26.04 & 45.95 & -4891.08 & 45.95 & -4891.08\\
\bottomrule\end{tabular}
\end{table*}

\begin{table*}[!t]
\caption{InternVL3-8B: Per-dataset scores (0--100) and end-to-end FLOPs savings (\%) relative to $F_{1024}$. $R$ denotes crop and $C$ whole-image compression; subscripts are visual-token caps. ZoomEye official final-cap and matched-F fallback are distinct configurations.}
\label{tab:comparators-internvl3-8b}
\centering\footnotesize\setlength{\tabcolsep}{3pt}
\begin{tabular}{cccccccccccccc}\toprule
 & & \multicolumn{2}{c}{\textbf{\boldmath $R_{1024}$}} & \multicolumn{2}{c}{\textbf{\boldmath $C_{256}$}} & \multicolumn{2}{c}{\textbf{\boldmath $C_{512}$}} & \multicolumn{2}{c}{\textbf{\boldmath CARES}} & \multicolumn{2}{c}{\textbf{\boldmath ZoomEye final-cap}} & \multicolumn{2}{c}{\textbf{\boldmath ZoomEye fallback}} \\
\textbf{\boldmath Dataset} & \textbf{\boldmath $N$} & \textbf{\boldmath Score} & \textbf{\boldmath Saved} & \textbf{\boldmath Score} & \textbf{\boldmath Saved} & \textbf{\boldmath Score} & \textbf{\boldmath Saved} & \textbf{\boldmath Score} & \textbf{\boldmath Saved} & \textbf{\boldmath Score} & \textbf{\boldmath Saved} & \textbf{\boldmath Score} & \textbf{\boldmath Saved} \\\midrule
AI2D & 1365 & 78.17 & -1.72 & 82.20 & 45.03 & 81.83 & 28.13 & 81.83 & 25.36 & 82.27 & -1656.13 & 82.27 & -1656.13\\
ChartQA & 13038 & 69.36 & 16.31 & 71.85 & 67.89 & 72.96 & 66.04 & 72.96 & 64.52 & 73.66 & -965.83 & 73.66 & -965.83\\
DocVQA & 17484 & 85.98 & -0.94 & 68.55 & 68.61 & 71.48 & 64.03 & 78.01 & 51.29 & 85.41 & -4467.92 & 85.40 & -4467.92\\
HR-Bench 8K & 376 & 57.45 & -12.76 & 44.15 & 54.74 & 50.53 & 27.92 & 49.73 & 24.65 & 74.20 & -21536.65 & 74.20 & -21536.65\\
InfoVQA & 1098 & 48.82 & 9.22 & 37.01 & 63.56 & 49.94 & 40.32 & 56.40 & 20.11 & 59.41 & -5435.88 & 59.41 & -5435.87\\
MME & 970 & 86.70 & -2.39 & 89.18 & 53.07 & 88.45 & 29.95 & 88.45 & 27.64 & 87.22 & -2974.21 & 87.22 & -2974.21\\
MMStar & 574 & 60.10 & -9.96 & 66.03 & 32.27 & 65.68 & 12.46 & 65.51 & 5.45 & 64.98 & -2706.66 & 64.98 & -2706.66\\
OCRBench & 404 & 62.38 & 2.32 & 71.53 & 59.73 & 68.07 & 39.36 & 67.82 & 32.71 & 70.30 & -1248.49 & 70.30 & -1248.49\\
Pinpoint & 192 & 27.60 & 7.60 & 28.65 & 63.19 & 28.12 & 40.33 & 28.12 & 17.39 & 49.48 & -13728.05 & 49.48 & -13728.05\\
POPE & 3618 & 88.42 & -13.25 & 89.19 & 24.17 & 89.25 & 5.01 & 89.25 & 1.26 & 89.72 & -1480.73 & 89.72 & -1480.73\\
RealWorldQA & 306 & 60.78 & -8.46 & 62.75 & 52.18 & 63.40 & 19.01 & 63.40 & 16.70 & 60.46 & -4141.35 & 60.46 & -4141.35\\
ScienceQA & 795 & 86.54 & -2.61 & 92.33 & 43.10 & 91.45 & 20.51 & 91.45 & 17.71 & 90.69 & -2079.66 & 90.69 & -2079.66\\
SEED-Bench Image & 5602 & 74.35 & 6.43 & 75.49 & 64.06 & 75.22 & 52.88 & 75.22 & 51.15 & 74.72 & -3043.13 & 74.72 & -3043.13\\
SOU & 7290 & 49.16 & -4.89 & 48.93 & 49.21 & 50.30 & 29.55 & 50.37 & 24.95 & 48.79 & -11729.89 & 48.79 & -11729.89\\
TextVQA & 15637 & 82.15 & 16.21 & 82.04 & 66.72 & 83.92 & 56.33 & 84.01 & 53.95 & 86.16 & -1889.00 & 86.16 & -1889.00\\
VizWiz & 13053 & 35.49 & 0.24 & 38.40 & 63.25 & 37.63 & 38.30 & 37.46 & 17.65 & 35.79 & -2593.70 & 35.79 & -2593.69\\
V$^*$ & 72 & 75.00 & 28.63 & 59.72 & 66.12 & 65.28 & 45.66 & 63.89 & 43.65 & 80.56 & -7937.79 & 80.56 & -7937.79\\
ZoomBench & 346 & 48.27 & 6.12 & 40.17 & 63.12 & 40.75 & 43.76 & 40.75 & 41.90 & 48.27 & -5405.70 & 48.27 & -5405.70\\
\bottomrule\end{tabular}
\end{table*}
\subsection{CropVLM comparison}
\label{sec:s-cropvlm}
CropVLM~\cite{cropvlm} constructs a learned crop and passes both the original image and crop to the target VLM. We cap each at 512 visual tokens, matching the total 1,024-token ceiling of the original-input baseline. Crop-constructor image longest edges of 512, 1,024, and 2,048 pixels are compared on the 7B validation set; the selected constructor is shared by all three targets. Constructor inference is included in end-to-end FLOPs, while final-answer tokens count the original image and crop. Each target is evaluated on all 82,220 test questions across 18 datasets. Table~\ref{tab:cropvlm-results} provides per-dataset results; Table~\ref{tab:s-inference-comparison} separates final-answer and total costs.
\begin{table*}[!t]
\caption{CropVLM results by dataset (\%). Savings are relative to each target's F1024.}
\label{tab:cropvlm-results}
\centering\scriptsize\setlength{\tabcolsep}{2pt}
\begin{threeparttable}
\begin{tabular*}{\textwidth}{@{\extracolsep{\fill}}cccccccccc@{}}\toprule
& \multicolumn{3}{c}{\textbf{\boldmath Qwen2.5-VL-7B}} & \multicolumn{3}{c}{\textbf{\boldmath Qwen2.5-VL-3B}} & \multicolumn{3}{c}{\textbf{\boldmath InternVL3-8B}}\\
\textbf{\boldmath Dataset} & \textbf{\boldmath Score} & \textbf{\boldmath \makecell[cc]{FLOPs\\saved}} & \textbf{\boldmath \makecell[cc]{Tokens\\saved}} & \textbf{\boldmath Score} & \textbf{\boldmath \makecell[cc]{FLOPs\\saved}} & \textbf{\boldmath \makecell[cc]{Tokens\\saved}} & \textbf{\boldmath Score} & \textbf{\boldmath \makecell[cc]{FLOPs\\saved}} & \textbf{\boldmath \makecell[cc]{Tokens\\saved}}\\\midrule
AI2D & 81.03 & -71.51 & -31.65 & 80.07 & -102.84 & -31.65 & 83.15 & -72.12 & -35.65 \\
ChartQA & 77.85 & -56.84 & -27.96 & 82.82 & -79.41 & -27.96 & 70.98 & 8.13 & 32.64 \\
DocVQA & 89.53 & -8.75 & 13.85 & 89.99 & -23.83 & 13.85 & 76.90 & -8.96 & 19.70 \\
HR-Bench 8K & 50.80 & -17.70 & 0.31 & 53.46 & -30.08 & 0.31 & 50.27 & -62.40 & -35.66 \\
InfoVQA & 63.81 & -13.55 & 1.31 & 57.97 & -23.63 & 1.31 & 52.65 & -33.18 & -14.22 \\
MME & 86.80 & -50.48 & -22.11 & 85.15 & -71.40 & -22.11 & 87.53 & -66.44 & -35.22 \\
MMStar & 62.20 & -82.00 & -33.54 & 56.27 & -124.60 & -33.54 & 66.55 & -104.67 & -71.11 \\
OCRBench & 71.78 & -17.39 & 15.10 & 77.72 & -39.93 & 15.10 & 69.31 & -44.20 & -19.77 \\
Pinpoint-Bench & 23.44 & -14.06 & 3.84 & 23.44 & -26.17 & 3.84 & 28.12 & -39.32 & -13.24 \\
POPE & 83.25 & -126.79 & -77.75 & 86.84 & -166.93 & -77.75 & 89.50 & -137.08 & -92.71 \\
RealWorldQA & 62.09 & -21.71 & -3.04 & 56.21 & -34.64 & -3.04 & 64.38 & -83.18 & -56.25 \\
ScienceQA & 82.39 & -59.62 & -3.96 & 74.34 & -101.48 & -3.96 & 91.70 & -81.50 & -48.00 \\
SEED-Bench Image & 75.58 & -23.58 & -3.58 & 73.51 & -37.64 & -3.58 & 75.72 & -11.92 & 14.76 \\
SOU & 52.67 & -22.91 & -2.27 & 51.80 & -38.44 & -2.27 & 49.77 & -48.42 & -25.99 \\
TextVQA & 75.10 & -12.07 & 8.23 & 78.71 & -25.71 & 8.23 & 84.50 & -10.86 & 15.05 \\
VizWiz & 44.03 & -21.92 & -2.23 & 46.66 & -35.50 & -2.23 & 35.93 & -44.50 & -18.80 \\
V$^*$ & 59.72 & -12.13 & 5.22 & 56.94 & -23.83 & 5.22 & 66.67 & -24.20 & -1.59 \\
ZoomBench & 35.26 & -13.77 & 4.34 & 37.86 & -26.02 & 4.34 & 39.31 & -29.37 & -4.48 \\
\midrule
Dataset macro & 65.41 & -35.93 & -8.66 & 64.99 & -56.23 & -8.66 & 65.72 & -49.68 & -21.70 \\
\bottomrule\end{tabular*}
\begin{tablenotes}[flushleft]\fontsize{7.5}{8.5}\selectfont\raggedright\setlength{\labelsep}{0pt}\item[] FLOPs include the crop constructor and final answer. The constructor image longest edge was selected on 7B validation: 512, 1,024, and 2,048 pixels scored 64.33, 64.37, and 64.45, respectively; the 2,048-pixel setting was frozen across targets.\end{tablenotes}
\end{threeparttable}
\end{table*}

\subsection{Performance--compute tradeoffs}
\label{sec:s-tradeoff}
\begin{figure}[t]
\centering
\includegraphics[width=\linewidth]{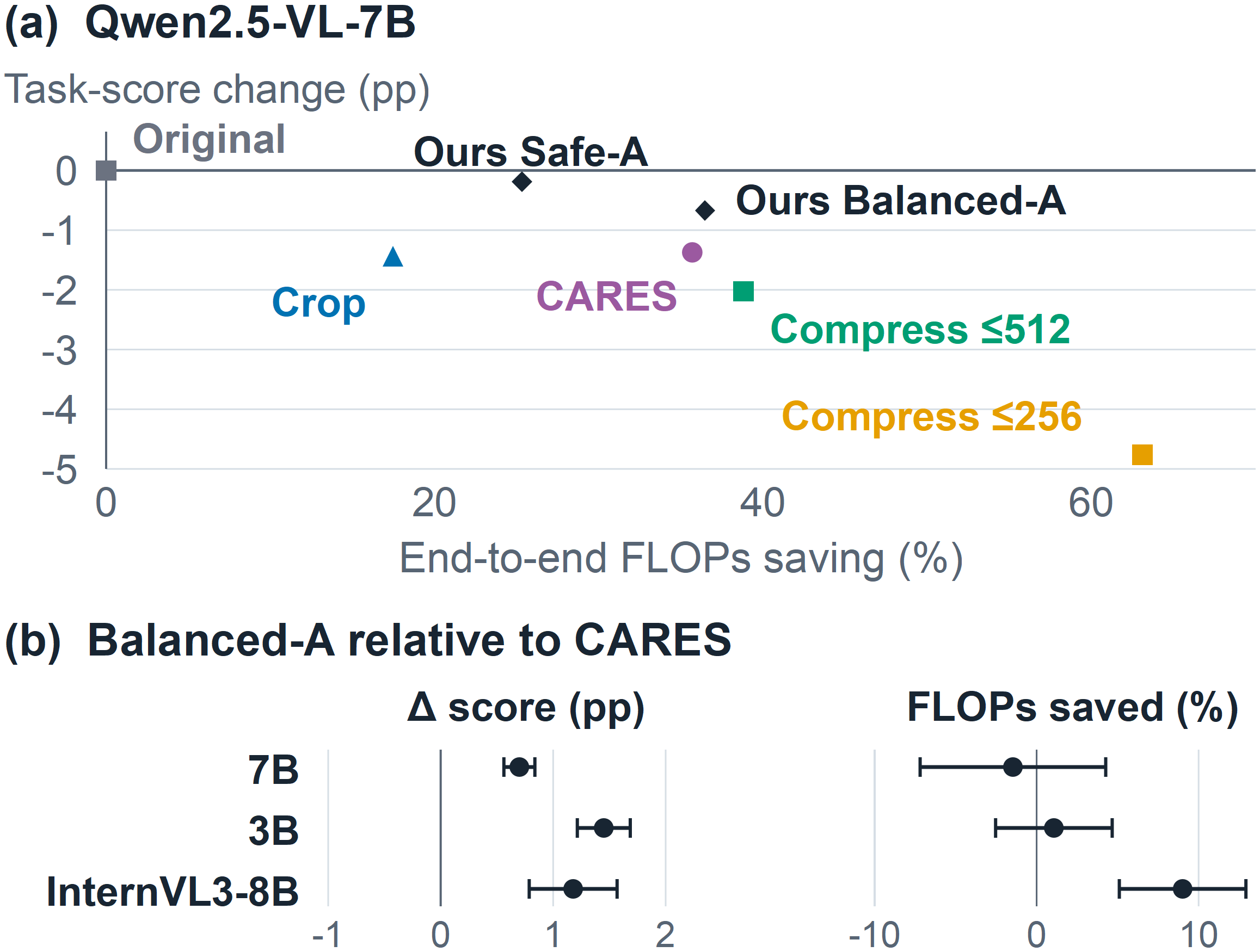}
\caption{Performance--compute tradeoffs (top) and Balanced-A versus CARES-F1024 (bottom). Ours: four-seed means; error bars: $\pm$1 standard deviation.}
\label{fig:points}
\end{figure}

Fig.~\ref{fig:points} visualizes the main table's accuracy--computation tradeoffs and the comparison with CARES-F1024.
\subsection{Cross-dataset stability}
\label{sec:s-stability}
Table~\ref{tab:s-stability} reports the mean, best, worst, and population standard deviation of score changes across the 18 datasets. For Ours, each dataset's change is first averaged over four seeds. This measures variation across datasets, rather than across training runs.
\begin{table*}[!t]
\caption{\textbf{Performance variation across 18 datasets.} Changes are relative to each target's own $F_{1024}$ baseline, in percentage points. Best and worst are the maximum and minimum signed score changes; SD is the population standard deviation across the 18 dataset-level changes. N/A denotes not evaluated. $\dagger$ marks InternVL3-8B Safe-M, feasible in 2/4 seeds.}
\label{tab:s-stability}
\centering
\fontsize{8}{9.5}\selectfont\setlength{\tabcolsep}{3pt}
\begin{threeparttable}
\begin{tabular*}{\textwidth}{@{\extracolsep{\fill}}ccccccc}\toprule
\textbf{\boldmath Method} & \textbf{\boldmath Mean $\Delta$} & \textbf{\boldmath Best $\Delta$} & \textbf{\boldmath Best dataset} & \textbf{\boldmath Worst $\Delta$} & \textbf{\boldmath Worst dataset} & \textbf{\boldmath SD} \\\midrule
\multicolumn{7}{c}{\textbf{Qwen2.5-VL-7B}}\\
CropVLM-F1024~\cite{cropvlm} & -1.62 & +2.27 & MME & -7.36 & InfoVQA & 2.79 \\
CARES-F1024~\cite{cares} & -1.37 & +2.09 & MMStar & -5.40 & InfoVQA & 2.36 \\
ViCrop 256~\cite{vicrop} & -3.56 & +2.16 & MME & -16.16 & InfoVQA & 4.85 \\
ViCrop 1024 & +0.97 & +7.80 & ZoomBench & -2.02 & TextVQA & 2.33 \\
ZoomEye~\cite{zoomeye} & +2.07 & +13.54 & Pinpoint-Bench & -1.04 & TextVQA & 4.17 \\
Ours Safe-M & -0.23 & +2.78 & V$^*$ & -1.69 & Pinpoint-Bench & 1.06 \\
Ours Safe-A & -0.19 & +3.13 & V$^*$ & -2.21 & Pinpoint-Bench & 1.15 \\
Ours Balanced-M & -0.91 & +1.18 & OCRBench & -3.57 & TextVQA & 1.51 \\
Ours Balanced-A & -0.67 & +4.17 & V$^*$ & -3.46 & DocVQA & 1.81 \\
\midrule
\multicolumn{7}{c}{\textbf{Qwen2.5-VL-3B}}\\
CropVLM-F1024~\cite{cropvlm} & -1.61 & +4.79 & HR-Bench 8K & -10.14 & InfoVQA & 3.33 \\
CARES-F1024~\cite{cares} & -2.27 & +0.72 & MME & -11.11 & V$^*$ & 2.72 \\
ViCrop 256~\cite{vicrop} & -4.04 & +0.98 & RealWorldQA & -23.14 & DocVQA & 7.02 \\
ViCrop 1024 & +0.49 & +3.12 & Pinpoint-Bench & -1.39 & V$^*$ & 1.12 \\
ZoomEye~\cite{zoomeye} & +2.20 & +16.67 & Pinpoint-Bench & -4.83 & DocVQA & 6.55 \\
Ours Safe-M & +0.10 & +4.12 & HR-Bench 8K & -1.17 & Pinpoint-Bench & 1.41 \\
Ours Safe-A & -0.00 & +2.93 & HR-Bench 8K & -1.30 & Pinpoint-Bench & 1.13 \\
Ours Balanced-M & -0.89 & +3.52 & HR-Bench 8K & -2.96 & ZoomBench & 1.65 \\
Ours Balanced-A & -0.81 & +3.92 & HR-Bench 8K & -3.03 & ZoomBench & 1.79 \\
\midrule
\multicolumn{7}{c}{\textbf{InternVL3-8B}}\\
CropVLM-F1024~\cite{cropvlm} & -1.17 & +4.46 & OCRBench & -14.93 & DocVQA & 4.41 \\
CARES-F1024~\cite{cares} & -1.08 & +2.97 & OCRBench & -13.83 & DocVQA & 3.68 \\
ViCrop 256~\cite{vicrop} & N/A & N/A & N/A & N/A & N/A & N/A \\
ViCrop 1024 & N/A & N/A & N/A & N/A & N/A & N/A \\
ZoomEye~\cite{zoomeye} & +3.23 & +22.92 & Pinpoint-Bench & -6.42 & DocVQA & 8.08 \\
Ours Safe-M$^{\dagger}$ & +0.86 & +9.65 & OCRBench & -1.99 & HR-Bench 8K & 2.38 \\
Ours Safe-A & +0.85 & +9.65 & OCRBench & -2.06 & HR-Bench 8K & 2.39 \\
Ours Balanced-M & +0.00 & +9.16 & OCRBench & -4.92 & HR-Bench 8K & 2.97 \\
Ours Balanced-A & +0.10 & +9.16 & OCRBench & -4.32 & HR-Bench 8K & 2.92 \\
\bottomrule\end{tabular*}
\begin{tablenotes}[flushleft]
\fontsize{7.5}{8.5}\selectfont\raggedright\setlength{\labelsep}{0pt}
\item[] ViCrop inputs both the original image and a crop. The same visual-token cap (256 or 1024) applies to each image.
\end{tablenotes}
\end{threeparttable}
\end{table*}
\subsection{Generalization to held-out datasets}
\label{sec:s-heldout}
We divide the 18 datasets into three folds of six. For each fold, we retrain the router and select its configuration using only the other 12 datasets, then evaluate on the held-out datasets' original test splits. Shared source groups belonging to a held-out dataset are excluded from training and validation. All learned upstream modules are retrained; only frozen input features, constructor outputs, and target outcomes are reused. The experiment covers 3 target models and four training seeds. Each dataset is tested once per seed, and Table~\ref{tab:domain-holdout} averages the resulting held-out scores.

The folds are: (1) V*, ZoomBench, ChartQA, RealWorldQA, TextVQA, and VizWiz; (2) Pinpoint-Bench, SOU, AI2D, ScienceQA, DocVQA, and MMStar; (3) HR-Bench 8K, MME, OCRBench, InfoVQA, SEED-Bench Image, and POPE. Balanced-A saves 30.87--35.26\% of end-to-end FLOPs across targets, with mean score losses of 1.17--1.96 points. Routing therefore retains computational savings on unseen datasets, with larger accuracy losses than in the main shared-dataset evaluation.
\begin{table*}[!t]
\caption{Held-out-domain results with both safety gates (mean over four seeds).}
\label{tab:domain-holdout}
\centering\scriptsize\setlength{\tabcolsep}{3pt}
\begin{tabular}{cccccccccc}\toprule
& & \multicolumn{4}{c}{\textbf{\boldmath Dataset macro}} & \multicolumn{4}{c}{\textbf{\boldmath Mechanism macro}}\\
\textbf{\boldmath Target} & \textbf{\boldmath Configuration} & \textbf{\boldmath Score (\%)} & \textbf{\boldmath $\Delta$ (pp)} & \textbf{\boldmath \makecell{FLOPs saved\\(\%)}} & \textbf{\boldmath \makecell{Tokens saved\\(\%)}} & \textbf{\boldmath Score (\%)} & \textbf{\boldmath $\Delta$ (pp)} & \textbf{\boldmath \makecell{FLOPs saved\\(\%)}} & \textbf{\boldmath \makecell{Tokens saved\\(\%)}}\\\midrule
Qwen2.5-VL-7B & Balanced-A & $65.12$ & $-1.91$ & $30.87$ & $35.94$ & $65.63$ & $-1.98$ & $31.42$ & $36.51$ \\
& Balanced-M & $65.19$ & $-1.84$ & $31.76$ & $36.68$ & $65.73$ & $-1.88$ & $32.20$ & $37.10$ \\
& Safe-A & $66.01$ & $-1.01$ & $17.03$ & $22.12$ & $66.56$ & $-1.05$ & $17.44$ & $22.59$ \\
& Safe-M & $66.17$ & $-0.85$ & $17.63$ & $21.65$ & $66.73$ & $-0.88$ & $18.09$ & $22.15$ \\
\midrule
Qwen2.5-VL-3B & Balanced-A & $64.63$ & $-1.96$ & $33.82$ & $38.44$ & $65.26$ & $-2.05$ & $33.69$ & $38.28$ \\
& Balanced-M & $64.64$ & $-1.96$ & $33.40$ & $38.13$ & $65.27$ & $-2.04$ & $33.29$ & $38.00$ \\
& Safe-A & $66.04$ & $-0.56$ & $9.60$ & $16.74$ & $66.70$ & $-0.61$ & $9.48$ & $16.61$ \\
& Safe-M & $65.80$ & $-0.80$ & $12.21$ & $18.75$ & $66.48$ & $-0.83$ & $12.01$ & $18.56$ \\
\midrule
InternVL3-8B & Balanced-A & $65.72$ & $-1.17$ & $35.26$ & $41.28$ & $66.06$ & $-1.28$ & $35.63$ & $41.61$ \\
& Balanced-M & $65.71$ & $-1.18$ & $35.34$ & $41.35$ & $66.06$ & $-1.28$ & $35.71$ & $41.67$ \\
& Safe-A & $66.39$ & $-0.49$ & $22.37$ & $27.93$ & $66.83$ & $-0.51$ & $22.50$ & $28.04$ \\
& Safe-M & $66.38$ & $-0.51$ & $22.55$ & $28.07$ & $66.82$ & $-0.52$ & $22.69$ & $28.18$ \\
\bottomrule\end{tabular}
\end{table*}
\section{Complete Ablation Results}
\subsection{Cropping and compression branches}
\label{sec:s-whole-branch}
The whole-branch ablation uses seed-0 weights and thresholds. Removing cropping forces the compression branch and its safety gate; removing compression uses cropping with its gate and original-input fallback. Table~\ref{tab:s-whole-branch} reports all configurations. Table~\ref{tab:s-branch-groups} reports Balanced-A by task group, averaging datasets equally within each group.
\begin{table*}[!t]
\caption{Whole-branch ablations across all configurations. Score changes and FLOPs savings are relative to each target's original input.}
\label{tab:s-whole-branch}
\centering\scriptsize\setlength{\tabcolsep}{3pt}
\begin{tabular}{ccccccccccc}\toprule
& & \multicolumn{3}{c}{\textbf{7B}} & \multicolumn{3}{c}{\textbf{3B}} & \multicolumn{3}{c}{\textbf{InternVL3-8B}}\\
\cmidrule(lr){3-5}\cmidrule(lr){6-8}\cmidrule(lr){9-11}
\textbf{Setting} & \textbf{Branches} & \textbf{Validation} & \textbf{$\Delta$ score (pp)} & \textbf{FLOPs (\%)} & \textbf{Validation} & \textbf{$\Delta$ score (pp)} & \textbf{FLOPs (\%)} & \textbf{Validation} & \textbf{$\Delta$ score (pp)} & \textbf{FLOPs (\%)}\\\midrule
Safe-M & Both & Yes & -0.313 & 28.96 & Yes & +0.097 & 21.21 & No & +1.019 & 25.11\\
Safe-M & Compression & No & -0.922 & 28.86 & No & -0.487 & 23.92 & No & +0.611 & 25.81\\
Safe-M & Crop & No & +0.927 & 3.90 & No & +0.638 & -1.43 & Yes & +0.611 & -4.58\\
Safe-A & Both & Yes & -0.115 & 27.64 & Yes & -0.111 & 21.67 & Yes & +0.849 & 25.47\\
Safe-A & Compression & No & -0.922 & 28.86 & Yes & -0.429 & 21.26 & No & +0.611 & 25.81\\
Safe-A & Crop & No & +0.927 & 3.90 & No & +0.638 & -1.43 & Yes & +0.611 & -4.58\\
Balanced-M & Both & Yes & -1.111 & 40.98 & Yes & -1.009 & 34.99 & Yes & +0.105 & 37.75\\
Balanced-M & Compression & No & -1.515 & 40.43 & Yes & -1.185 & 34.38 & Yes & -0.014 & 37.93\\
Balanced-M & Crop & Yes & +1.061 & 5.08 & Yes & +0.567 & -0.09 & Yes & +0.680 & -3.60\\
Balanced-A & Both & Yes & -0.535 & 39.09 & Yes & -0.862 & 35.18 & Yes & +0.104 & 37.75\\
Balanced-A & Compression & No & -1.994 & 39.81 & Yes & -1.185 & 34.38 & Yes & -0.014 & 37.93\\
Balanced-A & Crop & Yes & +0.867 & 9.01 & Yes & +0.567 & -0.09 & Yes & +0.680 & -3.60\\
\bottomrule\end{tabular}
\par\vspace{6pt}
\caption{Balanced-A branch ablations by task group.}
\label{tab:s-branch-groups}
\centering\small\setlength{\tabcolsep}{4pt}
\begin{tabular}{cccccccc}\toprule
\textbf{\boldmath Target} & \textbf{\boldmath Task group} & \multicolumn{2}{c}{\textbf{\boldmath Full router}} & \multicolumn{2}{c}{\textbf{\boldmath No crop}} & \multicolumn{2}{c}{\textbf{\boldmath No compression}}\\
 & & \textbf{\boldmath Score (\%)} & \textbf{\boldmath \makecell[cc]{FLOPs\\saved (\%)}} & \textbf{\boldmath Score (\%)} & \textbf{\boldmath \makecell[cc]{FLOPs\\saved (\%)}} & \textbf{\boldmath Score (\%)} & \textbf{\boldmath \makecell[cc]{FLOPs\\saved (\%)}}\\\midrule
Qwen2.5-VL-7B & Local detail & 47.70 & 44.72 & 43.14 & 47.17 & 51.55 & 12.46\\
Qwen2.5-VL-7B & Global context & 76.77 & 36.16 & 76.65 & 36.52 & 76.96 & 7.77\\
Qwen2.5-VL-7B & Text / document & 77.35 & 38.69 & 76.90 & 39.57 & 79.00 & 4.32\\
Qwen2.5-VL-7B & Mixed perception & 66.27 & 36.13 & 66.00 & 34.98 & 65.87 & 10.94\\
Qwen2.5-VL-3B & Local detail & 45.20 & 44.63 & 43.98 & 40.79 & 48.67 & 5.69\\
Qwen2.5-VL-3B & Global context & 75.19 & 42.68 & 75.18 & 43.14 & 75.69 & -2.13\\
Qwen2.5-VL-3B & Text / document & 79.85 & 10.02 & 80.03 & 10.84 & 81.01 & -6.90\\
Qwen2.5-VL-3B & Mixed perception & 65.48 & 39.17 & 65.37 & 38.94 & 65.79 & 2.04\\
InternVL3-8B & Local detail & 46.45 & 37.61 & 46.03 & 38.28 & 50.76 & -0.84\\
InternVL3-8B & Global context & 79.97 & 46.06 & 79.97 & 46.06 & 79.00 & -5.33\\
InternVL3-8B & Text / document & 76.48 & 33.24 & 76.48 & 33.24 & 76.02 & -2.14\\
InternVL3-8B & Mixed perception & 66.96 & 32.03 & 66.96 & 32.03 & 65.83 & -6.34\\
\bottomrule\end{tabular}

\end{table*}
\subsection{Safety gates}
\label{sec:s-safety}
For each trained router, we keep its proposal policy and thresholds fixed and toggle the two safety gates without retuning. Table~\ref{tab:s-safety-complete} reports all 48 conditions across the three targets, including four-seed means, sample standard deviations, and counts satisfying the validation constraints.
\begin{table*}[!t]
\caption{Complete matched safety ablations across three targets. Score and FLOPs savings are four-seed means $\pm$ sample SD; other values are means. Feasible counts are out of four.}
\label{tab:s-safety-complete}\label{tab:safety}\label{tab:safety-3b}\label{tab:safety-internvl3_8b}
\centering\scriptsize\setlength{\tabcolsep}{2pt}
\begin{tabular}{ccccccccccc}\toprule
 & & & \multicolumn{4}{c}{\textbf{\boldmath Dataset macro}} & \multicolumn{4}{c}{\textbf{\boldmath Mechanism macro}}\\
\cmidrule(lr){4-7}\cmidrule(lr){8-11}
\textbf{\boldmath Setting} & \textbf{\boldmath Safety} & \textbf{\boldmath Feasible} & \textbf{\boldmath Score} & \textbf{\boldmath $\Delta$pp} & \textbf{\boldmath FLOPs (\%)} & \textbf{\boldmath Tokens (\%)} & \textbf{\boldmath Score} & \textbf{\boldmath $\Delta$pp} & \textbf{\boldmath FLOPs (\%)} & \textbf{\boldmath Tokens (\%)}\\\midrule
\multicolumn{11}{c}{\textbf{Qwen2.5-VL-7B}}\\
Safe-M & None & 0/4 & 65.93 $\pm$ 0.27 & -1.092 & 42.29 $\pm$ 2.80 & 48.00 & 66.42 $\pm$ 0.29 & -1.194 & 42.33 $\pm$ 2.78 & 48.07\\
Safe-M & Crop & 0/4 & 65.75 $\pm$ 0.45 & -1.274 & 38.79 $\pm$ 2.48 & 44.30 & 66.27 $\pm$ 0.45 & -1.344 & 38.83 $\pm$ 2.41 & 44.36\\
Safe-M & Compression & 0/4 & 66.97 $\pm$ 0.14 & -0.052 & 28.85 $\pm$ 4.89 & 34.13 & 67.54 $\pm$ 0.12 & -0.075 & 28.42 $\pm$ 4.98 & 33.73\\
Safe-M & Both & 4/4 & 66.79 $\pm$ 0.33 & -0.234 & 25.35 $\pm$ 5.24 & 30.42 & 67.39 $\pm$ 0.30 & -0.225 & 24.92 $\pm$ 5.36 & 30.02\\
Safe-A & None & 0/4 & 65.81 $\pm$ 0.22 & -1.211 & 41.96 $\pm$ 3.09 & 47.80 & 66.30 $\pm$ 0.25 & -1.313 & 42.03 $\pm$ 3.04 & 47.90\\
Safe-A & Crop & 0/4 & 65.74 $\pm$ 0.39 & -1.281 & 38.29 $\pm$ 2.95 & 43.91 & 66.26 $\pm$ 0.40 & -1.355 & 38.36 $\pm$ 2.87 & 44.00\\
Safe-A & Compression & 0/4 & 66.91 $\pm$ 0.13 & -0.119 & 28.99 $\pm$ 4.46 & 34.42 & 67.47 $\pm$ 0.12 & -0.142 & 28.54 $\pm$ 4.58 & 34.00\\
Safe-A & Both & 4/4 & 66.84 $\pm$ 0.18 & -0.189 & 25.32 $\pm$ 5.24 & 30.52 & 67.43 $\pm$ 0.16 & -0.184 & 24.87 $\pm$ 5.34 & 30.10\\
Balanced-M & None & 0/4 & 65.48 $\pm$ 0.48 & -1.542 & 44.66 $\pm$ 6.05 & 49.85 & 65.98 $\pm$ 0.49 & -1.628 & 44.67 $\pm$ 6.02 & 49.86\\
Balanced-M & Crop & 0/4 & 65.31 $\pm$ 0.54 & -1.711 & 43.50 $\pm$ 6.44 & 48.63 & 65.83 $\pm$ 0.54 & -1.780 & 43.53 $\pm$ 6.40 & 48.66\\
Balanced-M & Compression & 4/4 & 66.28 $\pm$ 0.40 & -0.741 & 38.55 $\pm$ 3.61 & 43.56 & 66.86 $\pm$ 0.36 & -0.750 & 38.18 $\pm$ 3.37 & 43.20\\
Balanced-M & Both & 4/4 & 66.11 $\pm$ 0.46 & -0.911 & 37.39 $\pm$ 4.13 & 42.34 & 66.71 $\pm$ 0.42 & -0.903 & 37.04 $\pm$ 3.87 & 41.99\\
Balanced-A & None & 0/4 & 65.82 $\pm$ 0.36 & -1.206 & 42.33 $\pm$ 5.15 & 47.72 & 66.31 $\pm$ 0.41 & -1.302 & 42.38 $\pm$ 5.19 & 47.79\\
Balanced-A & Crop & 0/4 & 65.73 $\pm$ 0.39 & -1.292 & 40.93 $\pm$ 5.44 & 46.24 & 66.24 $\pm$ 0.42 & -1.374 & 41.01 $\pm$ 5.45 & 46.34\\
Balanced-A & Compression & 4/4 & 66.44 $\pm$ 0.06 & -0.587 & 37.87 $\pm$ 3.32 & 43.14 & 66.99 $\pm$ 0.07 & -0.619 & 37.62 $\pm$ 3.22 & 42.91\\
Balanced-A & Both & 4/4 & 66.35 $\pm$ 0.14 & -0.673 & 36.47 $\pm$ 3.69 & 41.67 & 66.92 $\pm$ 0.11 & -0.692 & 36.25 $\pm$ 3.56 & 41.46\\
\midrule
\multicolumn{11}{c}{\textbf{Qwen2.5-VL-3B}}\\
Safe-M & None & 0/4 & 65.09 $\pm$ 0.15 & -1.508 & 37.88 $\pm$ 2.11 & 45.56 & 65.58 $\pm$ 0.14 & -1.725 & 37.94 $\pm$ 2.03 & 45.70\\
Safe-M & Crop & 0/4 & 65.44 $\pm$ 0.24 & -1.161 & 33.82 $\pm$ 3.31 & 41.31 & 65.96 $\pm$ 0.23 & -1.351 & 33.76 $\pm$ 3.17 & 41.34\\
Safe-M & Compression & 1/4 & 66.35 $\pm$ 0.15 & -0.248 & 25.58 $\pm$ 2.58 & 33.16 & 66.98 $\pm$ 0.14 & -0.330 & 24.98 $\pm$ 2.65 & 32.65\\
Safe-M & Both & 4/4 & 66.70 $\pm$ 0.22 & +0.099 & 21.52 $\pm$ 2.14 & 28.92 & 67.35 $\pm$ 0.20 & +0.044 & 20.80 $\pm$ 2.27 & 28.29\\
Safe-A & None & 0/4 & 64.84 $\pm$ 0.59 & -1.752 & 39.88 $\pm$ 5.06 & 46.90 & 65.37 $\pm$ 0.57 & -1.935 & 39.89 $\pm$ 4.98 & 46.99\\
Safe-A & Crop & 0/4 & 65.04 $\pm$ 0.74 & -1.561 & 36.42 $\pm$ 6.52 & 43.29 & 65.59 $\pm$ 0.72 & -1.725 & 36.35 $\pm$ 6.42 & 43.29\\
Safe-A & Compression & 3/4 & 66.40 $\pm$ 0.31 & -0.193 & 22.81 $\pm$ 7.15 & 29.65 & 67.07 $\pm$ 0.33 & -0.244 & 22.23 $\pm$ 7.05 & 29.16\\
Safe-A & Both & 4/4 & 66.59 $\pm$ 0.20 & -0.002 & 19.35 $\pm$ 7.66 & 26.03 & 67.28 $\pm$ 0.21 & -0.034 & 18.69 $\pm$ 7.52 & 25.46\\
Balanced-M & None & 0/4 & 64.47 $\pm$ 0.53 & -2.128 & 43.43 $\pm$ 3.86 & 49.44 & 65.02 $\pm$ 0.50 & -2.294 & 43.35 $\pm$ 3.81 & 49.42\\
Balanced-M & Crop & 0/4 & 64.55 $\pm$ 0.62 & -2.042 & 42.57 $\pm$ 4.23 & 48.54 & 65.10 $\pm$ 0.59 & -2.206 & 42.47 $\pm$ 4.18 & 48.49\\
Balanced-M & Compression & 3/4 & 65.62 $\pm$ 0.18 & -0.973 & 36.07 $\pm$ 1.56 & 42.03 & 66.28 $\pm$ 0.22 & -1.029 & 35.34 $\pm$ 1.94 & 41.35\\
Balanced-M & Both & 4/4 & 65.71 $\pm$ 0.14 & -0.887 & 35.21 $\pm$ 1.40 & 41.13 & 66.37 $\pm$ 0.17 & -0.941 & 34.45 $\pm$ 1.75 & 40.42\\
Balanced-A & None & 0/4 & 64.61 $\pm$ 0.35 & -1.986 & 43.51 $\pm$ 3.74 & 49.59 & 65.14 $\pm$ 0.34 & -2.166 & 43.43 $\pm$ 3.71 & 49.57\\
Balanced-A & Crop & 0/4 & 64.66 $\pm$ 0.46 & -1.933 & 42.50 $\pm$ 3.81 & 48.54 & 65.20 $\pm$ 0.44 & -2.108 & 42.40 $\pm$ 3.78 & 48.49\\
Balanced-A & Compression & 3/4 & 65.73 $\pm$ 0.32 & -0.868 & 36.33 $\pm$ 1.76 & 42.36 & 66.37 $\pm$ 0.36 & -0.936 & 35.58 $\pm$ 2.09 & 41.67\\
Balanced-A & Both & 4/4 & 65.78 $\pm$ 0.24 & -0.815 & 35.31 $\pm$ 1.83 & 41.30 & 66.43 $\pm$ 0.27 & -0.878 & 34.55 $\pm$ 2.13 & 40.59\\
\midrule
\multicolumn{11}{c}{\textbf{InternVL3-8B}}\\
Safe-M & None & 0/4 & 66.06 $\pm$ 0.55 & -0.834 & 40.92 $\pm$ 1.68 & 47.68 & 66.30 $\pm$ 0.54 & -1.039 & 41.03 $\pm$ 1.67 & 47.71\\
Safe-M & Crop & 0/4 & 66.03 $\pm$ 0.59 & -0.864 & 40.66 $\pm$ 1.71 & 47.40 & 66.29 $\pm$ 0.60 & -1.050 & 40.78 $\pm$ 1.70 & 47.44\\
Safe-M & Compression & 1/4 & 67.78 $\pm$ 0.54 & +0.894 & 25.64 $\pm$ 7.37 & 31.11 & 68.25 $\pm$ 0.54 & +0.906 & 25.39 $\pm$ 7.34 & 30.78\\
Safe-M & Both & 2/4 & 67.75 $\pm$ 0.53 & +0.864 & 25.38 $\pm$ 7.42 & 30.84 & 68.24 $\pm$ 0.53 & +0.895 & 25.14 $\pm$ 7.39 & 30.51\\
Safe-A & None & 0/4 & 65.91 $\pm$ 0.56 & -0.980 & 40.88 $\pm$ 1.91 & 47.66 & 66.17 $\pm$ 0.58 & -1.169 & 40.95 $\pm$ 1.93 & 47.64\\
Safe-A & Crop & 0/4 & 65.96 $\pm$ 0.81 & -0.931 & 40.62 $\pm$ 2.18 & 47.39 & 66.24 $\pm$ 0.86 & -1.102 & 40.69 $\pm$ 2.22 & 47.37\\
Safe-A & Compression & 2/4 & 67.69 $\pm$ 0.42 & +0.800 & 25.51 $\pm$ 7.33 & 31.00 & 68.16 $\pm$ 0.43 & +0.815 & 25.28 $\pm$ 7.30 & 30.68\\
Safe-A & Both & 3/4 & 67.74 $\pm$ 0.47 & +0.849 & 25.26 $\pm$ 7.41 & 30.73 & 68.22 $\pm$ 0.47 & +0.882 & 25.02 $\pm$ 7.37 & 30.40\\
Balanced-M & None & 0/4 & 65.18 $\pm$ 0.23 & -1.712 & 42.94 $\pm$ 2.65 & 49.52 & 65.46 $\pm$ 0.23 & -1.884 & 43.01 $\pm$ 2.58 & 49.52\\
Balanced-M & Crop & 0/4 & 65.18 $\pm$ 0.23 & -1.711 & 42.94 $\pm$ 2.65 & 49.52 & 65.46 $\pm$ 0.23 & -1.883 & 43.00 $\pm$ 2.58 & 49.52\\
Balanced-M & Compression & 3/4 & 66.89 $\pm$ 0.25 & +0.001 & 37.72 $\pm$ 2.86 & 43.96 & 67.37 $\pm$ 0.23 & +0.027 & 37.32 $\pm$ 2.72 & 43.46\\
Balanced-M & Both & 4/4 & 66.89 $\pm$ 0.25 & +0.002 & 37.72 $\pm$ 2.86 & 43.96 & 67.37 $\pm$ 0.23 & +0.028 & 37.32 $\pm$ 2.72 & 43.46\\
Balanced-A & None & 0/4 & 65.33 $\pm$ 0.47 & -1.556 & 42.76 $\pm$ 2.92 & 49.37 & 65.60 $\pm$ 0.44 & -1.744 & 42.84 $\pm$ 2.82 & 49.39\\
Balanced-A & Crop & 0/4 & 65.32 $\pm$ 0.44 & -1.571 & 42.72 $\pm$ 2.98 & 49.34 & 65.59 $\pm$ 0.41 & -1.756 & 42.81 $\pm$ 2.87 & 49.35\\
Balanced-A & Compression & 3/4 & 67.01 $\pm$ 0.42 & +0.116 & 37.84 $\pm$ 2.60 & 44.13 & 67.47 $\pm$ 0.38 & +0.130 & 37.43 $\pm$ 2.49 & 43.62\\
Balanced-A & Both & 4/4 & 66.99 $\pm$ 0.39 & +0.101 & 37.81 $\pm$ 2.66 & 44.10 & 67.46 $\pm$ 0.36 & +0.117 & 37.40 $\pm$ 2.54 & 43.59\\
\bottomrule\end{tabular}
\end{table*}
\section{Budget and Resource Results}
\subsection{Visual-token-cap comparison}
\label{sec:s-budget-reference}
We compare Qwen2.5-VL-7B~\cite{qwen25vl} at caps of 16,384 and 1,024 visual tokens on 672 paired questions: all 72 V* test questions plus one question from each of 200 image/document groups in DocVQA, RealWorldQA, and AI2D. Seeded hashes (20261025) select groups and questions without using scores.

The comparison keeps the model, BF16 precision, SDPA attention, prompts, decoding, and evaluator fixed. Only the maximum input budget changes; the minimum remains 256 visual tokens. All calls complete on RTX 3090 at batch size 1. The main 18-dataset Qwen experiments use eager attention, while both conditions in this paired study use SDPA.

Table~\ref{tab:s-budget-reference} reports score changes and savings, with 95\% intervals from 5,000 paired natural-group bootstrap resamples. Table~\ref{tab:s-budget-resources} gives the corresponding absolute costs and peak memory. The cap saves 55.64\% FLOPs and 54.32\% tokens with a 0.63-point mean score loss; peak allocated memory decreases from 18.81 to 15.75 GiB.
\begin{table*}[!t]
\caption{\textbf{Paired visual-token-cap comparison: Qwen2.5-VL-7B.} Both conditions use SDPA and differ only in the maximum input budget. Scores use each dataset's metric on a 0--100 scale; $\Delta$ is the 1,024-cap score minus the 16,384-cap score. Savings use the higher-budget condition as the denominator within each dataset. Mechanism macro equally weights local detail (V*), text/document (DocVQA), and global context (RealWorldQA and AI2D). All 672 pairs complete; intervals use 5,000 paired natural-group bootstrap resamples.}
\label{tab:s-budget-reference}
\centering\small\setlength{\tabcolsep}{4pt}
\begin{tabular}{ccccccc}\toprule
\textbf{\boldmath Dataset} & \textbf{\boldmath Questions} & \textbf{\boldmath 16,384-cap score} & \textbf{\boldmath 1,024-cap score} & \textbf{\boldmath $\Delta$ score (pp)} & \textbf{\boldmath FLOPs saved (\%)} & \textbf{\boldmath Tokens saved (\%)}\\
\midrule
V* & 72 & 75.00 & 70.83 & -4.17 & 79.71 & 77.63\\
DocVQA & 200 & 98.94 & 99.10 & +0.16 & 80.62 & 78.06\\
RealWorldQA & 200 & 68.00 & 69.50 & +1.50 & 46.48 & 45.80\\
AI2D & 200 & 79.50 & 79.50 & +0.00 & 15.76 & 15.78\\
\midrule
Dataset macro & 672 & 80.36 & 79.73 & -0.63 & 55.64 & 54.32\\
Mechanism macro & 672 & 82.56 & 81.48 & -1.09 & 63.81 & 62.16\\
\bottomrule\end{tabular}
\medskip
\begin{tabular}{ccc}\toprule
\textbf{\boldmath Paired 95\% interval} & \textbf{\boldmath $\Delta$ score (pp)} & \textbf{\boldmath FLOPs saved (\%)}\\
\midrule
V* & $[-14.93,+6.67]$ & $[78.97,80.57]$\\
DocVQA & $[-0.74,+1.44]$ & $[79.26,81.98]$\\
RealWorldQA & $[-2.76,+5.88]$ & $[45.17,47.73]$\\
AI2D & $[-1.48,+1.49]$ & $[9.51,21.79]$\\
\bottomrule\end{tabular}
\end{table*}

\begin{table*}[!t]
\caption{\textbf{Absolute costs in the paired 7B SDPA study.} Visual tokens and measured operator FLOPs are per-question means; the final two rows average these quantities equally across datasets. Memory is the maximum PyTorch allocated/reserved memory observed among the corresponding calls, with the final rows taking the maximum across datasets. Batch size is one on an RTX 3090. This study uses one direct target call per condition; no routing frontend is executed.}
\label{tab:s-budget-resources}
\centering\small\setlength{\tabcolsep}{7pt}
\begin{tabular}{cccccc}\toprule
\textbf{\boldmath Dataset} & \textbf{\boldmath Token cap} & \textbf{\boldmath Visual tokens} & \textbf{\boldmath TFLOPs/query} & \textbf{\boldmath Allocated GiB} & \textbf{\boldmath Reserved GiB}\\
\midrule
V* & 16,384 & 4499.25 & 98.01 & 17.07 & 18.71\\
V* & 1,024 & 1006.49 & 19.89 & 15.74 & 16.13\\
\midrule
DocVQA & 16,384 & 4456.81 & 99.09 & 18.81 & 22.04\\
DocVQA & 1,024 & 977.99 & 19.21 & 15.74 & 16.13\\
\midrule
RealWorldQA & 16,384 & 1724.49 & 34.65 & 16.01 & 16.62\\
RealWorldQA & 1,024 & 934.64 & 18.55 & 15.75 & 16.13\\
\midrule
AI2D & 16,384 & 512.05 & 10.53 & 16.11 & 16.82\\
AI2D & 1,024 & 431.24 & 8.87 & 15.75 & 16.13\\
\midrule
Macro / peak & 16,384 & 2798.15 & 60.57 & 18.81 & 22.04\\
Macro / peak & 1,024 & 837.59 & 16.63 & 15.75 & 16.13\\
\bottomrule\end{tabular}
\end{table*}
\subsection{Computation and tokens}
\label{sec:s-costs}
Table~\ref{tab:cost} reports absolute stage costs and shares for Ours and CARES; Table~\ref{tab:search-cost} reports localization and search costs for the cropping comparisons. Stage shares pool test questions across runs and divide by the corresponding policy's total end-to-end FLOPs.

Table~\ref{tab:s-inference-comparison} separates final-answer inference from the complete execution path. Final-answer costs include vision encoding, projection, prefill, and decode. All target-call costs additionally include localization and search calls; end-to-end costs also include the router frontend. ViCrop's final visual tokens include both the original image and its crop. CARES~\cite{cares} reports prefill savings in its original Table~2, while the final-answer costs here also include decode.

Savings are computed from summed costs or tokens within each dataset before averaging datasets or task groups. Absolute TFLOPs/query and token counts use dataset-macro means, whereas stage shares use pooled totals. Thus the denominators differ between the savings and stage-breakdown tables.
\begin{table*}[!t]
\caption{\textbf{Per-stage costs of Ours Balanced-A and CARES-F1024.} TFLOPs/query is pooled FLOPs divided by 82,220 test questions and $10^{12}$; each share divides the stage total by the same candidate's end-to-end total. For Ours, the four final-answer rows are alternative executed inputs. Small nonzero head costs are shown in scientific notation.}\label{tab:cost}
\centering
\footnotesize\setlength{\tabcolsep}{2.5pt}
\begin{tabular}{ccccccc}\toprule
 & \multicolumn{2}{c}{\textbf{\boldmath 7B}} & \multicolumn{2}{c}{\textbf{\boldmath 3B}} & \multicolumn{2}{c}{\textbf{\boldmath InternVL3-8B}}\\
\textbf{\boldmath Stage} & \textbf{\boldmath TFLOPs/query} & \textbf{\boldmath Share (\%)} & \textbf{\boldmath TFLOPs/query} & \textbf{\boldmath Share (\%)} & \textbf{\boldmath TFLOPs/query} & \textbf{\boldmath Share (\%)} \\\midrule
\multicolumn{7}{c}{\textbf{Ours Balanced-A}}\\
Shared features & 0.258 & 2.60 & 0.258 & 3.95 & 0.258 & 3.04 \\
Selection head & $4.95\times10^{-7}$ & $<0.01$ & $4.95\times10^{-7}$ & $<0.01$ & $4.95\times10^{-7}$ & $<0.01$ \\
Crop constructor & 0.068 & 0.68 & 0.047 & 0.71 & 0.001 & 0.01 \\
Resolution head & $4.51\times10^{-7}$ & $<0.01$ & $4.64\times10^{-7}$ & $<0.01$ & $4.92\times10^{-7}$ & $<0.01$ \\
Degradation features & 0 & 0 & 0 & 0 & 0 & 0 \\
Crop risk head & $6.66\times10^{-10}$ & $<0.01$ & $4.54\times10^{-10}$ & $<0.01$ & $1.13\times10^{-11}$ & $<0.01$ \\
Compression risk head & $4.52\times10^{-7}$ & $<0.01$ & $4.66\times10^{-7}$ & $<0.01$ & $4.94\times10^{-7}$ & $<0.01$ \\
Final answer: Crop & 0.415 & 4.19 & 0.154 & 2.36 & 0.010 & 0.12 \\
Final answer: Compress 256 & 0.692 & 6.99 & 0.573 & 8.78 & 2.211 & 26.13 \\
Final answer: Compress 512 & 6.797 & 68.68 & 2.796 & 42.88 & 3.221 & 38.07 \\
Final answer: Original & 1.669 & 16.86 & 2.695 & 41.32 & 2.761 & 32.63 \\
Total & 9.897 & 100.00 & 6.522 & 100.00 & 8.462 & 100.00 \\
\midrule\multicolumn{7}{c}{\textbf{CARES-F1024}}\\
Shared features & 0.258 & 2.72 & 0.258 & 4.46 & 0.258 & 3.15 \\
Resolution head & $4.93\times10^{-7}$ & $<0.01$ & $4.93\times10^{-7}$ & $<0.01$ & $4.93\times10^{-7}$ & $<0.01$ \\
Final answer & 9.211 & 97.28 & 5.512 & 95.54 & 7.925 & 96.85 \\
Total & 9.468 & 100.00 & 5.770 & 100.00 & 8.183 & 100.00 \\
\bottomrule\end{tabular}
\end{table*}

\begin{table*}[!t]
\caption{\textbf{Localization and search breakdown.} Each number is a percentage of the candidate's pooled end-to-end FLOPs. ViCrop jointly uses the original image and a crop; ZoomEye official final-cap and matched-F fallback are separate comparisons; the latter includes its fallback answer cost in the final-answer column.}\label{tab:search-cost}
\centering
\footnotesize\setlength{\tabcolsep}{4pt}
\begin{threeparttable}
\begin{tabular}{ccccccc}\toprule
\textbf{\boldmath Target} & \textbf{\boldmath Method} & \textbf{\boldmath General attn.} & \textbf{\boldmath Question attn.} & \textbf{\boldmath Cue generation/filter} & \textbf{\boldmath Search} & \textbf{\boldmath Final answer}\\\midrule
7B & ViCrop 256~\cite{vicrop} & 30.01 & 31.90 & 0 & 0 & 38.09 \\
7B & ViCrop 1024 & 32.54 & 33.18 & 0 & 0 & 34.28 \\
3B & ViCrop 256 & 30.26 & 31.74 & 0 & 0 & 38.00 \\
3B & ViCrop 1024 & 32.65 & 33.14 & 0 & 0 & 34.20 \\
7B & ZoomEye official final-cap~\cite{zoomeye} & 0 & 0 & 4.42 & 87.25 & 8.34 \\
7B & ZoomEye matched-F fallback & 0 & 0 & 4.41 & 87.19 & 8.39 \\
3B & ZoomEye official final-cap~\cite{zoomeye} & 0 & 0 & 1.09 & 96.79 & 2.11 \\
3B & ZoomEye matched-F fallback & 0 & 0 & 1.09 & 96.79 & 2.11 \\
InternVL3-8B & ZoomEye official final-cap~\cite{zoomeye} & 0 & 0 & 1.87 & 95.61 & 2.52 \\
InternVL3-8B & ZoomEye matched-F fallback & 0 & 0 & 1.87 & 95.61 & 2.52 \\
\bottomrule\end{tabular}
\begin{tablenotes}[flushleft]
\fontsize{7.5}{8.5}\selectfont\raggedright\setlength{\labelsep}{0pt}
\item[] ViCrop inputs both the original image and a crop. The same visual-token cap (256 or 1024) applies to each image.
\end{tablenotes}
\end{threeparttable}
\end{table*}

\begin{table*}[!t]
\caption{Final-answer and end-to-end costs across 18 datasets. D / M denotes dataset-macro / mechanism-balanced macro. TFLOPs/query and visual-token counts are dataset-macro means; savings are relative to each target's original input. ZoomEye official final-cap and matched-F fallback are distinct configurations. N/A indicates an unevaluated method.}
\label{tab:s-inference-comparison}\label{tab:inference-7b}\label{tab:inference-3b}\label{tab:inference-internvl38b}
\centering\footnotesize\setlength{\tabcolsep}{2pt}
\begin{threeparttable}
\begin{tabular}{cccccccc}\toprule
 & \textbf{\boldmath $\Delta$score} & \multicolumn{2}{c}{\textbf{\boldmath Final-answer inference}} & \multicolumn{3}{c}{\textbf{\boldmath Saving vs. own original input (\%)}} & \textbf{\boldmath End-to-end}\\
\textbf{\boldmath Method} & \textbf{\boldmath (pp; D / M)} & \textbf{\boldmath TFLOPs/query} & \textbf{\boldmath Vis. tokens} & \textbf{\boldmath \makecell[cc]{Final-answer\\FLOPs (D / M)}} & \textbf{\boldmath Vis. tokens (D / M)} & \textbf{\boldmath \makecell[cc]{End-to-end\\FLOPs (D / M)}} & \textbf{\boldmath TFLOPs/query}\\\midrule
\multicolumn{8}{c}{\textbf{Qwen2.5-VL-7B}}\\
Original & 0.00 / 0.00 & 15.17 & 757.1 & 0.00 / 0.00 & 0.00 / 0.00 & 0.00 / 0.00 & 15.17 \\
Crop & -1.43 / -1.71 & 11.96 & 586.0 & 22.70 / 22.72 & 24.50 / 24.51 & 17.46 / 17.48 & 12.65 \\
Compress 256 & -4.77 / -4.93 & 4.99 & 224.0 & 63.10 / 62.93 & 66.49 / 66.27 & 63.10 / 62.93 & 4.99 \\
Compress 512 & -2.02 / -2.01 & 8.69 & 421.2 & 38.81 / 38.71 & 40.47 / 40.34 & 38.81 / 38.71 & 8.69 \\
Ours Balanced-A & -0.67 / -0.69 & 9.03 & 437.1 & 39.60 / 39.42 & 41.67 / 41.46 & 36.47 / 36.25 & 9.41 \\
ViCrop 256~\cite{vicrop} & -3.56 / -3.77 & 6.23 & 289.8 & 53.69 / 53.48 & 56.33 / 56.06 & -20.65 / -21.22 & 16.32 \\
ViCrop 1024 & 0.97 / 0.84 & 16.12 & 805.3 & -5.43 / -5.46 & -5.40 / -5.43 & -203.10 / -203.13 & 46.86 \\
CARES-F1024~\cite{cares} & -1.37 / -1.39 & 9.13 & 443.8 & 37.69 / 37.75 & 39.36 / 39.39 & 35.69 / 35.75 & 9.40 \\
CropVLM-F1024~\cite{cropvlm} & -1.62 / -1.68 & 15.63 & 785.5 & -7.53 / -7.62 & -8.66 / -8.78 & -35.93 / -36.17 & 19.33 \\
ZoomEye official final-cap~\cite{zoomeye} & 2.07 / 1.86 & 14.11 & 695.6 & 8.26 / 8.17 & 9.68 / 9.58 & -2328.90 / -2182.36 & 449.20 \\
ZoomEye matched-F fallback & 1.96 / 1.75 & 14.47 & 716.4 & 4.00 / 3.88 & 4.62 / 4.48 & -2333.16 / -2186.65 & 449.56 \\
\midrule
\multicolumn{8}{c}{\textbf{Qwen2.5-VL-3B}}\\
Original & 0.00 / 0.00 & 8.91 & 757.1 & 0.00 / 0.00 & 0.00 / 0.00 & 0.00 / 0.00 & 8.91 \\
Crop & -1.97 / -2.23 & 6.98 & 586.0 & 23.30 / 23.32 & 24.50 / 24.51 & 14.29 / 14.30 & 7.67 \\
Compress 256 & -5.82 / -6.14 & 2.80 & 224.0 & 64.50 / 64.32 & 66.49 / 66.27 & 64.50 / 64.32 & 2.80 \\
Compress 512 & -2.00 / -2.12 & 5.00 & 421.2 & 39.80 / 39.69 & 40.47 / 40.34 & 39.80 / 39.69 & 5.00 \\
Ours Balanced-A & -0.81 / -0.88 & 5.26 & 440.6 & 40.20 / 39.52 & 41.30 / 40.59 & 35.31 / 34.55 & 5.61 \\
ViCrop 256~\cite{vicrop} & -4.04 / -4.38 & 3.53 & 289.5 & 54.96 / 54.73 & 56.31 / 56.04 & -17.47 / -18.08 & 9.25 \\
ViCrop 1024 & 0.49 / 0.44 & 9.45 & 803.5 & -5.24 / -5.28 & -5.20 / -5.24 & -203.26 / -203.30 & 27.54 \\
CARES-F1024~\cite{cares} & -2.27 / -2.28 & 5.50 & 463.6 & 36.46 / 36.37 & 37.12 / 37.01 & 33.03 / 32.93 & 5.76 \\
CropVLM-F1024~\cite{cropvlm} & -1.61 / -1.73 & 9.13 & 785.5 & -7.37 / -7.47 & -8.66 / -8.78 & -56.23 / -56.58 & 12.84 \\
ZoomEye official final-cap~\cite{zoomeye} & 2.20 / 1.81 & 7.59 & 637.0 & 15.14 / 15.26 & 16.35 / 16.45 & -4603.66 / -4436.00 & 480.03 \\
ZoomEye matched-F fallback & 2.21 / 1.82 & 7.59 & 637.5 & 15.03 / 15.14 & 16.22 / 16.33 & -4603.77 / -4436.11 & 480.04 \\
\midrule
\multicolumn{8}{c}{\textbf{InternVL3-8B}}\\
Original & 0.00 / 0.00 & 12.90 & 710.7 & 0.00 / 0.00 & 0.00 / 0.00 & 0.00 / 0.00 & 12.90 \\
Crop & -1.52 / -1.78 & 11.73 & 637.8 & 7.71 / 7.62 & 8.56 / 8.43 & 2.01 / 1.93 & 12.42 \\
Compress 256 & -3.10 / -3.33 & 5.41 & 256.0 & 55.56 / 55.53 & 61.21 / 61.09 & 55.56 / 55.53 & 5.41 \\
Compress 512 & -1.65 / -1.86 & 7.75 & 399.3 & 36.64 / 36.84 & 39.93 / 40.08 & 36.64 / 36.84 & 7.75 \\
Ours Balanced-A & 0.10 / 0.12 & 7.57 & 387.8 & 40.02 / 39.61 & 44.10 / 43.59 & 37.81 / 37.40 & 7.84 \\
CARES-F1024~\cite{cares} & -1.08 / -1.19 & 8.39 & 438.0 & 32.04 / 31.97 & 34.98 / 34.85 & 29.89 / 29.82 & 8.65 \\
CropVLM-F1024~\cite{cropvlm} & -1.17 / -1.31 & 14.63 & 813.0 & -19.16 / -19.06 & -21.70 / -21.62 & -49.68 / -49.65 & 18.33 \\
ZoomEye official final-cap~\cite{zoomeye} & 3.23 / 2.84 & 12.82 & 692.9 & -0.51 / 0.08 & 0.95 / 1.56 & -5278.93 / -5036.86 & 716.34 \\
ZoomEye matched-F fallback & 3.23 / 2.84 & 12.82 & 692.9 & -0.51 / 0.08 & 0.95 / 1.56 & -5278.93 / -5036.86 & 716.34 \\
ViCrop 256~\cite{vicrop} & \multicolumn{7}{c}{N/A} \\
ViCrop 1024 & \multicolumn{7}{c}{N/A} \\
\bottomrule\end{tabular}
\begin{tablenotes}[flushleft]
\fontsize{7.5}{8.5}\selectfont\raggedright\setlength{\labelsep}{0pt}
\item[] ViCrop inputs both the original image and a crop. The same visual-token cap (256 or 1024) applies to each image.
\end{tablenotes}
\end{threeparttable}
\end{table*}
\subsection{GPU memory}
\label{sec:s-pipeline-memory}
\noindent\textbf{Budget stress test.}\quad
On 36 inputs spanning all 18 datasets, each target retains its main attention backend and generation settings. The caps are 1,024 versus 16,384 tokens for Qwen, and four versus twelve tiles (1,024 versus 3,072 tokens) for InternVL3. Inputs combine seeded samples and the largest images. At batch size 1, with the allocator cache cleared before each condition, the larger cap causes 22/36 and 25/36 out-of-memory failures on 3B and 7B; all capped inputs complete. Table~\ref{tab:s-budget-memory} separates these failures from memory peaks among successful calls.

\noindent\textbf{Complete pipeline.}\quad
Original-input inference and the router pipeline run in fresh processes on the same RTX 3090 using the same 100 inputs across all 18 datasets. The pipeline loads both frontend and target on the GPU, interleaves all four configurations without clearing the cache, and probes each reduced route and branch fallback twice. Table~\ref{tab:pipeline-memory} reports maximum allocated/reserved memory over loading, natural requests, and probes at batch size 1. All three targets complete these runs.

The reported visual-token budgets are caps. Qwen2.5-VL~\cite{qwen25vl} aligns processed dimensions to multiples of 28; InternVL3~\cite{internvl3} uses 256 tokens per tile. Our InternVL3 original-input setting permits at most four tiles without a whole-image thumbnail; the two compression settings permit one or at most two tiles. The cap, preprocessing, and evaluation split can therefore produce different scores from published native-resolution evaluations.
\begin{table*}[!t]
\caption{\textbf{Input-budget memory stress test under the main attention settings.} All targets use the same 36 inputs spanning 18 datasets, at batch size one on an RTX 3090. Allocated/reserved peaks summarize successful calls; failed calls are counted in the out-of-memory column. The higher-budget Qwen peaks therefore cover fewer inputs and do not estimate the memory required to complete the full set. The 1,024-token cap completes all 36 inputs for every target.}
\label{tab:s-budget-memory}
\centering\small\setlength{\tabcolsep}{7pt}
\begin{tabular}{ccccc}\toprule
\textbf{\boldmath Target} & \textbf{\boldmath Visual-token cap} & \textbf{\boldmath Out of memory / attempted} & \textbf{\boldmath Allocated GiB} & \textbf{\boldmath Reserved GiB}\\
\midrule
Qwen2.5-VL-3B & 16,384 & 22 / 36 & 16.01 & 18.37\\
Qwen2.5-VL-3B & 1,024 & 0 / 36 & 9.70 & 10.64\\
\midrule
Qwen2.5-VL-7B & 16,384 & 25 / 36 & 18.07 & 18.96\\
Qwen2.5-VL-7B & 1,024 & 0 / 36 & 18.09 & 18.96\\
\midrule
InternVL3-8B & 3,072 & 0 / 36 & 17.88 & 19.23\\
InternVL3-8B & 1,024 & 0 / 36 & 15.34 & 15.61\\
\bottomrule\end{tabular}
\medskip
\caption{Peak PyTorch GPU memory at batch size 1. Each own-$F_{1024}$ and resident-pipeline run uses the same 100 inputs spanning all 18 datasets. Values are observed maxima including model loading, natural routes, and physical branch probes, rather than sums of separate stage peaks.}
\label{tab:pipeline-memory}
\centering\footnotesize\setlength{\tabcolsep}{7pt}
\begin{tabular}{ccccc}\toprule
 & \multicolumn{2}{c}{\textbf{\boldmath Own-$F_{1024}$ (GiB)}} & \multicolumn{2}{c}{\textbf{\boldmath Resident pipeline (GiB)}}\\
\cmidrule(lr){2-3}\cmidrule(lr){4-5}
\textbf{\boldmath Target} & \textbf{\boldmath Allocated} & \textbf{\boldmath Reserved} & \textbf{\boldmath Allocated} & \textbf{\boldmath Reserved}\\\midrule
Qwen2.5-VL-7B & 18.09 & 22.68 & 19.43 & 22.66 \\
Qwen2.5-VL-3B & 9.70 & 18.28 & 11.01 & 18.02 \\
InternVL3-8B & 15.51 & 15.77 & 17.46 & 17.90 \\
\bottomrule\end{tabular}
\end{table*}

\subsection{Latency}
\label{sec:s-latency}
We compare $F_{1024}$, CARES-$F_{1024}$, and Balanced-A on Qwen2.5-VL-7B using one RTX 3090 at batch size 1, with BF16 precision and the main eager-attention setting. Seeded hashes select one test question from each of 10 natural image/document groups per dataset, giving 180 questions across all 18 datasets. The sample selection does not use answers or routing decisions. Balanced-A uses the frozen router with both safety gates; CARES uses its continuous resolution selector.

All methods run with the same prompt and generation settings for each dataset. The target and frontend models remain loaded on the GPU, and each request recomputes its frontend features. After six warmup requests per method and an untimed crop pass to initialize the localizer, we measure three repetitions with balanced method order, yielding 1,620 timed requests. GPU synchronization brackets frontend execution and the target call. Timing includes image/question preprocessing, online routing and localization when selected, and answer generation; model loading and disk image decoding are excluded. Continuous GPU monitoring records no other compute or graphics processes during the measurements.

Table~\ref{tab:s-latency} reports aggregate and per-dataset results. We first average repetitions per question and questions per dataset. Overall times average the 18 dataset means; overall savings average the relative savings within each dataset. Confidence intervals use 5,000 paired bootstrap resamples of natural image/document groups. Balanced-A reduces latency on 13 of the 18 datasets. Its target-model stage is slightly faster than CARES, but higher frontend overhead increases total latency. High-resolution inputs are particularly affected: HR-Bench 8K uses compression for all sampled questions yet has substantial frontend time, indicating that latency overhead is not confined to localization.
\begin{table*}[!t]
\caption{Controlled latency on Qwen2.5-VL-7B. Times are in milliseconds; savings are relative to $F_{1024}$, with negative values indicating slower execution.}
\label{tab:s-latency}
\centering\footnotesize\setlength{\tabcolsep}{4pt}
\begin{tabular}{cccccc}
\toprule
\multicolumn{6}{c}{\textbf{(a) Aggregate latency}} \\
\textbf{Method} & \textbf{Frontend (ms)} & \textbf{Target (ms)} & \textbf{E2E (ms)} & \textbf{Savings (\%)} & \textbf{95\% CI (\%)} \\
\midrule
$F_{1024}$ & 0.01 & 765.32 & 765.33 & -- & -- \\
CARES-$F_{1024}$ & 115.68 & 502.65 & 618.33 & 18.25 & [14.61, 21.36] \\
Ours Balanced-A & 167.45 & 491.53 & 658.98 & 13.72 & [9.47, 17.52] \\
\midrule
\multicolumn{6}{c}{\textbf{(b) Per-dataset latency}} \\
\textbf{Dataset} & \textbf{\makecell{F1024\\E2E}} & \textbf{\makecell{CARES\\E2E}} & \textbf{\makecell{Ours\\E2E}} & \textbf{\makecell{Ours\\frontend}} & \textbf{\makecell{Savings (\%)\\CARES / Ours}} \\
\midrule
AI2D & 417.24 & 395.96 & 365.08 & 48.60 & 5.10 / 12.50 \\
ChartQA & 583.44 & 469.98 & 537.16 & 66.43 & 19.45 / 7.93 \\
DocVQA & 948.16 & 618.69 & 678.00 & 103.31 & 34.75 / 28.49 \\
HR-Bench 8K & 1118.01 & 1309.10 & 1417.33 & 658.30 & -17.09 / -26.77 \\
InfoVQA & 840.07 & 696.02 & 852.58 & 155.56 & 17.15 / -1.49 \\
MME & 473.32 & 372.40 & 375.61 & 64.52 & 21.32 / 20.64 \\
MMStar & 423.36 & 385.13 & 394.63 & 58.03 & 9.03 / 6.79 \\
OCRBench & 1074.14 & 878.42 & 1111.48 & 134.42 & 18.22 / -3.48 \\
Pinpoint-Bench & 977.32 & 1142.19 & 1392.73 & 680.42 & -16.87 / -42.51 \\
POPE & 360.89 & 375.46 & 426.99 & 174.41 & -4.04 / -18.32 \\
RealWorldQA & 826.05 & 507.33 & 488.64 & 89.24 & 38.58 / 40.85 \\
ScienceQA & 638.16 & 460.88 & 438.72 & 42.87 & 27.78 / 31.25 \\
SEED-Bench Image & 1155.18 & 515.07 & 528.81 & 96.22 & 55.41 / 54.22 \\
SOU & 642.93 & 519.61 & 501.21 & 107.06 & 19.18 / 22.04 \\
TextVQA & 747.33 & 479.69 & 516.22 & 91.95 & 35.81 / 30.93 \\
VizWiz & 879.44 & 758.43 & 610.62 & 74.68 & 13.76 / 30.57 \\
V$^*$ & 831.45 & 592.20 & 616.37 & 240.79 & 28.78 / 25.87 \\
ZoomBench & 839.43 & 653.48 & 609.46 & 127.29 & 22.15 / 27.40 \\
\bottomrule
\end{tabular}
\end{table*}

\section{Combination with Visual Token Pruning}
\label{sec:s-pruning-combination}
\noindent\textbf{Direct combination.}\quad
We apply VisionZip~\cite{visionzip} after input routing, keeping the existing Balanced-A router from seed 0 fixed. Original-input and routed-image inference are evaluated at 100\%, 70\%, and 50\% token retention on all 82,220 test questions across 18 datasets. Matched inference reruns give scores of 67.04 for the unpruned original input and 66.49 for routing alone. Direct combination at 70\% retention saves 50.29\% of end-to-end FLOPs and 61.13\% of visual tokens with a 1.38-point score loss; at 50\%, savings reach 57.65\% and 72.27\%, with a 3.95-point loss. These results use one frozen router; the main standalone comparison averages four training seeds.
\begin{table*}[!t]
\caption{7B direct VisionZip composition by dataset. Score changes (pp) and savings (\%) use the paired unpruned original-input baseline.}
\label{tab:s-visionzip-detailed}
\label{tab:s-visionzip-direct50}
\centering\scriptsize\setlength{\tabcolsep}{3pt}
\begin{tabular}{ccccccccccc}\toprule
& & \multicolumn{3}{c}{\textbf{\boldmath Ours}} & \multicolumn{3}{c}{\textbf{\boldmath VisionZip}} & \multicolumn{3}{c}{\textbf{\boldmath Ours + VisionZip (direct)}}\\
\textbf{\boldmath Dataset} & \textbf{\boldmath Original score} & \textbf{\boldmath $\Delta$} & \textbf{\boldmath FLOPs} & \textbf{\boldmath Tokens} & \textbf{\boldmath $\Delta$} & \textbf{\boldmath FLOPs} & \textbf{\boldmath Tokens} & \textbf{\boldmath $\Delta$} & \textbf{\boldmath FLOPs} & \textbf{\boldmath Tokens}\\\midrule
\multicolumn{11}{c}{\textbf{70\% visual tokens retained}}\\
AI2D & 81.25 & +0.37 & 35.38 & 41.40 & +0.44 & 19.87 & 30.18 & +0.95 & 46.91 & 59.16 \\
ChartQA & 76.32 & -0.99 & 20.22 & 24.52 & -0.60 & 19.96 & 30.05 & -2.39 & 35.26 & 47.33 \\
DocVQA & 92.39 & -3.57 & 49.10 & 51.18 & -0.92 & 21.08 & 30.09 & -8.12 & 59.18 & 65.93 \\
HR-Bench 8K & 53.72 & -4.26 & 46.96 & 50.05 & +1.86 & 20.80 & 30.12 & -4.79 & 57.14 & 65.14 \\
InfoVQA & 70.98 & -2.66 & 25.86 & 27.68 & -3.44 & 21.08 & 30.09 & -9.27 & 40.96 & 49.47 \\
MME & 84.54 & -0.41 & 45.41 & 49.60 & -0.10 & 20.80 & 30.15 & +0.00 & 55.72 & 64.86 \\
MMStar & 60.63 & +1.39 & 17.71 & 27.79 & -1.92 & 18.29 & 30.25 & -1.05 & 30.91 & 49.74 \\
OCRBench & 74.75 & +0.99 & 38.80 & 46.62 & +1.24 & 20.22 & 30.16 & -1.73 & 49.52 & 62.81 \\
Pinpoint-Bench & 27.60 & -2.08 & 43.09 & 46.25 & -0.52 & 21.13 & 30.08 & -1.04 & 54.26 & 62.45 \\
POPE & 85.54 & -0.14 & 31.00 & 43.66 & -0.17 & 20.23 & 30.19 & -0.30 & 42.42 & 60.82 \\
RealWorldQA & 62.75 & -2.61 & 43.59 & 46.57 & -0.65 & 20.91 & 30.15 & +0.00 & 54.53 & 62.68 \\
ScienceQA & 82.89 & -0.50 & 36.23 & 47.14 & -0.13 & 19.01 & 30.26 & -0.13 & 46.31 & 63.30 \\
SEED-Bench Image & 76.53 & -1.27 & 49.61 & 52.77 & +0.04 & 20.74 & 30.09 & -1.37 & 59.21 & 67.03 \\
SOU & 53.31 & -1.25 & 40.07 & 48.11 & -0.22 & 18.81 & 30.11 & -1.28 & 49.63 & 63.75 \\
TextVQA & 79.97 & -3.53 & 41.00 & 43.85 & -0.18 & 21.11 & 30.10 & -3.87 & 52.66 & 60.79 \\
VizWiz & 42.28 & -0.05 & 46.21 & 48.32 & -0.10 & 21.12 & 30.15 & -0.52 & 56.87 & 63.91 \\
V$^*$ & 66.67 & +8.33 & 48.90 & 53.78 & +0.00 & 20.98 & 30.08 & +8.33 & 58.48 & 67.75 \\
ZoomBench & 34.68 & +2.31 & 44.58 & 47.57 & +0.29 & 20.87 & 30.09 & +1.73 & 55.31 & 63.40 \\
\midrule
Dataset mean & 67.04 & -0.55 & 39.09 & 44.27 & -0.28 & 20.39 & 30.13 & -1.38 & 50.29 & 61.13 \\
\midrule
Task-group mean & 67.63 & -0.61 & 38.93 & 44.12 & -0.32 & 20.40 & 30.13 & -1.57 & 50.16 & 61.02 \\
\bottomrule\end{tabular}
\end{table*}
\begin{table*}[!t]
\centering
{\small Table~\ref{tab:s-visionzip-detailed} (continued).}\par\vspace{5pt}
\scriptsize\setlength{\tabcolsep}{3pt}
\begin{tabular}{ccccccccccc}\toprule
& & \multicolumn{3}{c}{\textbf{\boldmath Ours}} & \multicolumn{3}{c}{\textbf{\boldmath VisionZip}} & \multicolumn{3}{c}{\textbf{\boldmath Ours + VisionZip (direct)}}\\
\textbf{\boldmath Dataset} & \textbf{\boldmath Original score} & \textbf{\boldmath $\Delta$} & \textbf{\boldmath FLOPs} & \textbf{\boldmath Tokens} & \textbf{\boldmath $\Delta$} & \textbf{\boldmath FLOPs} & \textbf{\boldmath Tokens} & \textbf{\boldmath $\Delta$} & \textbf{\boldmath FLOPs} & \textbf{\boldmath Tokens}\\\midrule
\multicolumn{11}{c}{\textbf{50\% visual tokens retained}}\\
AI2D & 81.25 & +0.37 & 35.38 & 41.40 & -0.07 & 32.91 & 50.18 & +0.15 & 54.47 & 70.87 \\
ChartQA & 76.32 & -0.99 & 20.22 & 24.52 & -2.38 & 33.13 & 50.04 & -5.02 & 45.12 & 62.41 \\
DocVQA & 92.39 & -3.57 & 49.10 & 51.18 & -5.01 & 34.91 & 50.10 & -18.91 & 65.82 & 75.69 \\
HR-Bench 8K & 53.72 & -4.26 & 46.96 & 50.05 & +1.60 & 34.40 & 50.12 & -5.05 & 63.84 & 75.12 \\
InfoVQA & 70.98 & -2.66 & 25.86 & 27.68 & -11.43 & 34.87 & 50.09 & -16.85 & 50.88 & 63.93 \\
MME & 84.54 & -0.41 & 45.41 & 49.60 & +0.21 & 34.43 & 50.16 & -0.41 & 62.49 & 74.93 \\
MMStar & 60.63 & +1.39 & 17.71 & 27.79 & -2.44 & 30.37 & 50.23 & -3.14 & 39.49 & 64.16 \\
OCRBench & 74.75 & +0.99 & 38.80 & 46.62 & -3.22 & 33.45 & 50.16 & -9.16 & 56.61 & 73.48 \\
Pinpoint-Bench & 27.60 & -2.08 & 43.09 & 46.25 & +1.56 & 34.98 & 50.10 & -3.13 & 61.60 & 73.20 \\
POPE & 85.54 & -0.14 & 31.00 & 43.66 & -0.03 & 33.60 & 50.27 & -1.46 & 49.88 & 72.07 \\
RealWorldQA & 62.75 & -2.61 & 43.59 & 46.57 & -0.33 & 34.58 & 50.14 & -1.31 & 61.74 & 73.39 \\
ScienceQA & 82.89 & -0.50 & 36.23 & 47.14 & +0.88 & 31.49 & 50.23 & -1.01 & 52.90 & 73.84 \\
SEED-Bench Image & 76.53 & -1.27 & 49.61 & 52.77 & -0.21 & 34.34 & 50.11 & -2.03 & 65.53 & 76.48 \\
SOU & 53.31 & -1.25 & 40.07 & 48.11 & -0.18 & 31.15 & 50.12 & -1.56 & 55.94 & 74.15 \\
TextVQA & 79.97 & -3.53 & 41.00 & 43.85 & -1.19 & 34.95 & 50.13 & -5.55 & 60.33 & 72.02 \\
VizWiz & 42.28 & -0.05 & 46.21 & 48.32 & -0.59 & 34.93 & 50.14 & -1.18 & 63.90 & 74.27 \\
V$^*$ & 66.67 & +8.33 & 48.90 & 53.78 & +1.39 & 34.73 & 50.10 & +4.17 & 64.74 & 76.98 \\
ZoomBench & 34.68 & +2.31 & 44.58 & 47.57 & +2.02 & 34.53 & 50.10 & +0.29 & 62.38 & 73.88 \\
\midrule
Dataset mean & 67.04 & -0.55 & 39.09 & 44.27 & -1.08 & 33.76 & 50.14 & -3.95 & 57.65 & 72.27 \\
\midrule
Task-group mean & 67.63 & -0.61 & 38.93 & 44.12 & -1.27 & 33.78 & 50.14 & -4.29 & 57.54 & 72.19 \\
\bottomrule\end{tabular}
\end{table*}

\noindent\textbf{Adapted combination and training-data control.}\quad
For each retention rate, we generate answers under $F_{1024}$, $R_{1024}$, $C_{256}$, and $C_{512}$ to retrain the frontend heads that depend on target outcomes. The target weights, image features, crop constructor, and head architectures remain fixed. A preselected pool contains 7,590 training and 3,701 validation questions, shared across all retention rates. It is sampled by natural-image/document groups within the existing splits. An unpruned router trained on this same pool controls for the change in training and validation data. Each trained router is evaluated on the complete 18-dataset test set.

Four-state labels pair the correctness of the original and reduced inputs at the same token retention rate $\rho$, in the order $(Y^{F_\rho},Y^{a_\rho})$. Thus, fallback restores the original image while retaining pruning. Reported savings use the matched unpruned original-input baseline. The original validation selection rule applies unchanged. Table~\ref{tab:s-visionzip-control} reports the matched-data comparison: the unpruned control loses 1.11 points with 39.41\% FLOPs savings, while the 70\% adapted combination changes score by +0.37 points with 29.44\% savings. At 50\%, the adapted combination loses 1.54 points with 36.58\% savings. Table~\ref{tab:s-visionzip-configurations} reports all M/A, Safe/Balanced, and four-gate configurations at each retention rate.
\begin{table*}[!t]
\caption{7B training-data control for adapting routing to VisionZip.}
\label{tab:s-visionzip-control}
\centering\footnotesize\setlength{\tabcolsep}{2pt}
\begin{threeparttable}
\begin{tabular*}{\linewidth}{@{\extracolsep{\fill}}ccccc@{}}\toprule
\textbf{\boldmath Method} & \textbf{\boldmath \makecell{Retained\\tokens (\%)}} & \textbf{\boldmath \makecell{$\Delta$score\\(pp)}} & \textbf{\boldmath \makecell{FLOPs\\saved (\%)}} & \textbf{\boldmath \makecell{Tokens\\saved (\%)}}\\\midrule
Same-data router & 100 & -1.11 & 39.41 & 43.97 \\
Adapted combination & 70 & +0.37 & 29.44 & 42.15 \\
Adapted combination & 50 & -1.54 & 36.58 & 54.92 \\
\bottomrule\end{tabular*}
\begin{tablenotes}[flushleft]
\fontsize{7.5}{8.5}\selectfont\raggedright\setlength{\labelsep}{0pt}
\item[] All heads use the same 7,590 training and 3,701 validation questions. Same-data router: heads trained using the unpruned target's answers. Adapted combination: heads trained using the pruned target's answers. Changes and savings use the unpruned original-input baseline.
\end{tablenotes}
\end{threeparttable}
\end{table*}

\begin{table*}[!t]
\caption{7B adapted VisionZip composition by dataset. Score changes (pp) and savings (\%) are relative to the unpruned original input.}
\label{tab:s-visionzip-adapted-detailed}
\centering\scriptsize\setlength{\tabcolsep}{3pt}
\begin{tabular}{ccccccccccc}\toprule
& & \multicolumn{3}{c}{\textbf{\boldmath Same-data router 100\%}} & \multicolumn{3}{c}{\textbf{\boldmath Adapted 70\%}} & \multicolumn{3}{c}{\textbf{\boldmath Adapted 50\%}}\\
\textbf{\boldmath Dataset} & \textbf{\boldmath Original score} & \textbf{\boldmath $\Delta$} & \textbf{\boldmath FLOPs} & \textbf{\boldmath Tokens} & \textbf{\boldmath $\Delta$} & \textbf{\boldmath FLOPs} & \textbf{\boldmath Tokens} & \textbf{\boldmath $\Delta$} & \textbf{\boldmath FLOPs} & \textbf{\boldmath Tokens}\\\midrule
AI2D & 81.25 & +0.29 & 47.64 & 54.64 & +0.95 & 37.07 & 49.81 & -0.29 & 31.43 & 51.37 \\
ChartQA & 76.32 & -3.11 & 43.98 & 49.67 & -0.66 & 15.91 & 30.17 & -2.58 & 32.21 & 51.15 \\
DocVQA & 92.39 & -3.84 & 29.73 & 31.76 & -0.92 & 19.79 & 30.09 & -5.02 & 33.65 & 50.12 \\
HR-Bench 8K & 53.72 & -4.26 & 44.45 & 47.46 & +6.65 & 29.27 & 39.53 & +1.60 & 33.00 & 50.12 \\
InfoVQA & 70.98 & -0.37 & 1.81 & 3.29 & -3.51 & 19.80 & 30.26 & -11.43 & 33.49 & 50.11 \\
MME & 84.54 & -1.24 & 61.34 & 66.03 & +0.31 & 47.75 & 57.52 & +0.21 & 36.96 & 54.20 \\
MMStar & 60.63 & +0.87 & 20.95 & 29.97 & -0.35 & 25.84 & 46.01 & -2.44 & 28.75 & 52.55 \\
OCRBench & 74.75 & +0.50 & 38.85 & 44.87 & +1.24 & 16.02 & 30.22 & -3.22 & 30.16 & 50.95 \\
Pinpoint-Bench & 27.60 & -2.60 & 42.64 & 44.73 & +1.04 & 24.46 & 35.66 & -2.08 & 43.82 & 58.26 \\
POPE & 85.54 & -0.88 & 27.39 & 33.07 & +0.33 & 31.10 & 53.37 & -0.14 & 42.96 & 69.23 \\
RealWorldQA & 62.75 & -1.31 & 49.51 & 52.90 & -0.33 & 33.59 & 44.39 & -1.96 & 48.85 & 62.91 \\
ScienceQA & 82.89 & -0.63 & 39.14 & 48.92 & -0.63 & 28.20 & 44.48 & +0.50 & 31.56 & 55.20 \\
SEED-Bench Image & 76.53 & -1.84 & 55.27 & 58.91 & -0.30 & 32.20 & 42.04 & -0.21 & 32.93 & 50.12 \\
SOU & 53.31 & +0.18 & 32.67 & 39.33 & +0.38 & 41.50 & 55.14 & +0.00 & 36.05 & 56.10 \\
TextVQA & 79.97 & -5.55 & 53.50 & 56.15 & -0.31 & 19.84 & 31.16 & -1.42 & 35.56 & 51.79 \\
VizWiz & 42.28 & +0.01 & 46.93 & 49.06 & -0.49 & 34.93 & 44.06 & -0.85 & 58.23 & 69.75 \\
V$^*$ & 66.67 & +4.17 & 29.04 & 33.70 & +4.17 & 40.79 & 53.12 & +1.39 & 31.37 & 51.18 \\
ZoomBench & 34.68 & -0.29 & 44.46 & 47.10 & -0.87 & 31.85 & 41.64 & +0.29 & 37.36 & 53.45 \\
\midrule
Dataset mean & 67.04 & -1.11 & 39.41 & 43.97 & +0.37 & 29.44 & 42.15 & -1.54 & 36.58 & 54.92 \\
\midrule
Task-group mean & 67.63 & -1.13 & 38.90 & 43.42 & +0.28 & 28.99 & 41.77 & -1.69 & 36.61 & 54.99 \\
\bottomrule\end{tabular}
\end{table*}

\noindent\textbf{Stronger pruning and fallback.}\quad
At 50\% retention, adapted Balanced-A reduces score loss by 2.42 points relative to direct combination, but sacrifices 21.07 percentage points of FLOPs savings. Compared with the unpruned matched-data control, adapted 50\% increases token savings from 43.97\% to 54.92\%, but FLOPs savings fall from 39.41\% to 36.58\% and score loss rises from 1.11 to 1.54 points. Its original-image fallback rate is 78.39\%, compared with 16.41\% for the unpruned matched-data control, averaged equally across datasets. Returning to the original image reduces input-processing savings while leaving pruning active. On InfoVQA and DocVQA, the adapted model falls back on 99.91\% and 99.94\% of samples, yet loses 11.43 and 5.02 points; pruning the original image alone loses 11.43 and 5.01 points. These cases show why fallback cannot remove information loss caused by pruning the original input. For adapted Balanced-A at 70\% and 50\%, the 95\% intervals for score change are $[-0.34,1.11]$ and $[-2.10,-0.96]$ points, respectively, from 5,000 paired bootstrap replicates over natural image/document groups.

\noindent\textbf{Cost and reporting.}\quad
End-to-end FLOPs include frontend computation and the measured pruned target call. Offline calls used to generate training labels are excluded from deployment cost. All comparisons use matched unpruned original-input inference as the reference. Table~\ref{tab:s-visionzip-detailed} reports direct combinations by dataset; Table~\ref{tab:s-visionzip-adapted-detailed} reports adapted combinations and the matched-data control. Both dataset and task-group means are included.
\begin{table*}[!t]
\caption{7B retrained routing configurations at three visual-token retention levels.}
\label{tab:s-visionzip-configurations}
\label{tab:s-visionzip-configurations-70}
\label{tab:s-visionzip-configurations-50}
\centering\fontsize{7.5}{8.5}\selectfont\renewcommand{\arraystretch}{0.95}\setlength{\tabcolsep}{4pt}
\begin{threeparttable}
\begin{tabular*}{\textwidth}{@{\extracolsep{\fill}}cccccccc@{}}\toprule
 & & \multicolumn{3}{c}{\textbf{\boldmath Dataset mean}} & \multicolumn{3}{c}{\textbf{\boldmath Task-group mean}}\\
\cmidrule(lr){3-5}\cmidrule(lr){6-8}
\textbf{\boldmath Config} & \textbf{\boldmath Gate} & \textbf{\boldmath $\Delta$score (pp)} & \textbf{\boldmath FLOPs saved (\%)} & \textbf{\boldmath Tokens saved (\%)} & \textbf{\boldmath $\Delta$score (pp)} & \textbf{\boldmath FLOPs saved (\%)} & \textbf{\boldmath Tokens saved (\%)}\\\midrule
\multicolumn{8}{c}{\textbf{100\% visual tokens retained}}\\
Safe-M & None & -0.822 & 32.99 & 40.36 & -1.012 & 33.35 & 40.71 \\
Safe-M & Crop & -0.238 & 26.44 & 33.31 & -0.367 & 26.79 & 33.67 \\
Safe-M & Compression & -0.062 & 25.69 & 32.85 & -0.178 & 25.37 & 32.53 \\
Safe-M & Both & +0.522 & 19.13 & 25.81 & +0.468 & 18.81 & 25.49 \\
Safe-A & None & -1.965 & 49.15 & 54.03 & -2.125 & 49.16 & 54.00 \\
Safe-A & Crop & -2.271 & 47.64 & 52.43 & -2.407 & 47.71 & 52.46 \\
Safe-A & Compression & +0.244 & 22.65 & 26.44 & +0.232 & 21.78 & 25.53 \\
Safe-A & Both & -0.062 & 21.14 & 24.85 & -0.050 & 20.33 & 23.99 \\
Balanced-M & None & -2.171 & 48.87 & 53.98 & -2.305 & 48.94 & 54.02 \\
Balanced-M & Crop & -2.368 & 48.12 & 53.19 & -2.485 & 48.25 & 53.30 \\
Balanced-M & Compression & -1.157 & 40.27 & 45.11 & -1.170 & 39.72 & 44.53 \\
Balanced-M & Both & -1.354 & 39.52 & 44.32 & -1.350 & 39.03 & 43.80 \\
Balanced-A & None & -1.965 & 49.15 & 54.03 & -2.125 & 49.16 & 54.00 \\
Balanced-A & Crop & -2.111 & 48.35 & 53.20 & -2.260 & 48.41 & 53.21 \\
Balanced-A & Compression & -0.960 & 40.20 & 44.81 & -0.999 & 39.64 & 44.20 \\
Balanced-A & Both & -1.105 & 39.41 & 43.97 & -1.134 & 38.90 & 43.42 \\
\bottomrule\end{tabular*}
\end{threeparttable}
\end{table*}
\begin{table*}[!t]
\centering
{\small Table~\ref{tab:s-visionzip-configurations} (continued).}\par\vspace{5pt}
\fontsize{7.5}{8.5}\selectfont\renewcommand{\arraystretch}{0.95}\setlength{\tabcolsep}{4pt}
\begin{threeparttable}
\begin{tabular*}{\textwidth}{@{\extracolsep{\fill}}cccccccc@{}}\toprule
 & & \multicolumn{3}{c}{\textbf{\boldmath Dataset mean}} & \multicolumn{3}{c}{\textbf{\boldmath Task-group mean}}\\
\cmidrule(lr){3-5}\cmidrule(lr){6-8}
\textbf{\boldmath Config} & \textbf{\boldmath Gate} & \textbf{\boldmath $\Delta$score (pp)} & \textbf{\boldmath FLOPs saved (\%)} & \textbf{\boldmath Tokens saved (\%)} & \textbf{\boldmath $\Delta$score (pp)} & \textbf{\boldmath FLOPs saved (\%)} & \textbf{\boldmath Tokens saved (\%)}\\\midrule
\multicolumn{8}{c}{\textbf{70\% visual tokens retained}}\\
Safe-M & None & -1.754 & 50.69 & 61.64 & -2.000 & 50.71 & 61.70 \\
Safe-M & Crop & -1.899 & 47.64 & 58.73 & -2.094 & 47.59 & 58.72 \\
Safe-M & Compression & +0.422 & 32.73 & 44.97 & +0.291 & 32.34 & 44.67 \\
Safe-M & Both & +0.277 & 29.67 & 42.06 & +0.198 & 29.22 & 41.70 \\
Safe-A & None & -1.647 & 50.77 & 61.59 & -1.891 & 50.77 & 61.62 \\
Safe-A & Crop & -1.732 & 48.11 & 59.07 & -1.951 & 48.09 & 59.08 \\
Safe-A & Compression & +0.352 & 32.35 & 44.51 & +0.241 & 31.96 & 44.19 \\
Safe-A & Both & +0.266 & 29.70 & 41.99 & +0.181 & 29.28 & 41.65 \\
Balanced-M & None & -1.754 & 50.69 & 61.64 & -2.000 & 50.71 & 61.70 \\
Balanced-M & Crop & -1.899 & 47.64 & 58.73 & -2.094 & 47.59 & 58.72 \\
Balanced-M & Compression & +0.422 & 32.73 & 44.97 & +0.291 & 32.34 & 44.67 \\
Balanced-M & Both & +0.277 & 29.67 & 42.06 & +0.198 & 29.22 & 41.70 \\
Balanced-A & None & -1.790 & 50.22 & 61.52 & -2.050 & 50.27 & 61.59 \\
Balanced-A & Crop & -1.632 & 46.28 & 57.77 & -1.844 & 46.30 & 57.81 \\
Balanced-A & Compression & +0.214 & 33.38 & 45.89 & +0.075 & 32.96 & 45.55 \\
Balanced-A & Both & +0.372 & 29.44 & 42.15 & +0.281 & 28.99 & 41.77 \\
\midrule
\multicolumn{8}{c}{\textbf{50\% visual tokens retained}}\\
Safe-M & None & -4.528 & 59.95 & 74.78 & -4.962 & 59.85 & 74.74 \\
Safe-M & Crop & -4.108 & 55.50 & 71.12 & -4.444 & 55.47 & 71.12 \\
Safe-M & Compression & -1.401 & 39.21 & 57.69 & -1.689 & 39.27 & 57.80 \\
Safe-M & Both & -0.980 & 34.75 & 54.03 & -1.171 & 34.89 & 54.19 \\
Safe-A & None & -4.332 & 60.22 & 74.95 & -4.774 & 60.12 & 74.90 \\
Safe-A & Crop & -3.950 & 56.00 & 71.46 & -4.297 & 55.97 & 71.47 \\
Safe-A & Compression & -1.251 & 39.09 & 57.55 & -1.545 & 39.15 & 57.66 \\
Safe-A & Both & -0.870 & 34.87 & 54.06 & -1.068 & 35.00 & 54.22 \\
Balanced-M & None & -5.525 & 57.25 & 71.32 & -5.776 & 57.26 & 71.35 \\
Balanced-M & Crop & -5.396 & 56.79 & 70.93 & -5.641 & 56.76 & 70.93 \\
Balanced-M & Compression & -1.686 & 36.92 & 54.72 & -1.847 & 36.98 & 54.81 \\
Balanced-M & Both & -1.558 & 36.46 & 54.34 & -1.711 & 36.49 & 54.40 \\
Balanced-A & None & -4.636 & 57.22 & 71.79 & -4.987 & 57.28 & 71.86 \\
Balanced-A & Crop & -4.615 & 54.87 & 69.86 & -4.946 & 55.01 & 70.00 \\
Balanced-A & Compression & -1.559 & 38.92 & 56.84 & -1.734 & 38.88 & 56.85 \\
Balanced-A & Both & -1.537 & 36.58 & 54.92 & -1.693 & 36.61 & 54.99 \\
\bottomrule\end{tabular*}
\begin{tablenotes}[flushleft]
\fontsize{7.5}{8.5}\selectfont\raggedright\setlength{\labelsep}{0pt}
\item[] All heads retrained on the same 7,590 training and 3,701 validation questions. Changes and savings are relative to the unpruned original input.
\end{tablenotes}
\end{threeparttable}
\end{table*}

\clearpage
\section{Supporting Evaluation Details}
\begin{table}[!ht]
\caption{Project split counts and evaluator identities. Training, validation, and test use the grouped splits described in the main paper.}\label{tab:splits}
\centering\setlength{\tabcolsep}{1.8pt}
\scriptsize
\begin{tabular}{ccccc}
\toprule
\textbf{\boldmath Dataset} & \textbf{\boldmath Training} & \textbf{\boldmath Validation} & \textbf{\boldmath Test} & \textbf{\boldmath Metric} \\
\midrule
AI2D~\cite{ai2d} & 1,105 & 618 & 1,365 & Multiple-choice accuracy \\
ChartQA~\cite{chartqa} & 12,984 & 6,677 & 13,038 & Relaxed accuracy \\
DocVQA~\cite{docvqa} & 19,291 & 8,037 & 17,484 & ANLS \\
HR-Bench 8K~\cite{hrbench} & 300 & 124 & 376 & Multiple-choice accuracy \\
InfoVQA~\cite{infovqa} & 1,110 & 593 & 1,098 & ANLS \\
MME~\cite{mme} & 938 & 466 & 970 & Yes/no accuracy \\
MMStar~\cite{mmstar} & 637 & 289 & 574 & Multiple-choice accuracy \\
OCRBench~\cite{ocrbench} & 394 & 201 & 404 & Substring matching \\
Pinpoint-Bench~\cite{pinpoint} & 171 & 70 & 192 & Semantic judge \\
POPE~\cite{pope} & 3,258 & 2,124 & 3,618 & Official scoring \\
RealWorldQA~\cite{realworldqa} & 333 & 126 & 306 & Exact match \\
ScienceQA~\cite{scienceqa} & 821 & 401 & 795 & Multiple-choice accuracy \\
SEED-Bench Image~\cite{seedbench} & 5,861 & 2,770 & 5,602 & Multiple-choice accuracy \\
SOU~\cite{sou} & 7,182 & 3,732 & 7,290 & Multiple-choice accuracy \\
TextVQA~\cite{textvqa} & 15,867 & 7,890 & 15,637 & VQA soft score \\
VizWiz~\cite{vizwiz} & 13,232 & 6,557 & 13,053 & VQA soft score \\
V$^*$~\cite{vstar} & 76 & 43 & 72 & Multiple-choice accuracy \\
ZoomBench~\cite{zoombench} & 342 & 156 & 346 & Official evaluator \\
\bottomrule\end{tabular}

\end{table}

\subsection{Validation configuration selection}
\label{sec:s-head-training}
Risk probabilities are calibrated on pooled validation data using a temperature and class-specific logit biases. Configuration selection searches compression thresholds $\{0.65,0.84\}$, crop-selection thresholds $\{0.2,0.3,\ldots,0.8\}$, and fallback tradeoffs $\{0,0.05,\ldots,2\}$, giving 574 combinations per M/A variant. Safe permits a dataset-macro score loss of 0.5 points and a worst-dataset loss of 2 points; Balanced permits 2 and 5 points.

With both gates enabled, we find the largest feasible FLOPs saving, then choose the highest mean score among candidates within 0.5 percentage points of that saving; ties prefer greater saving. If none meets both limits, selection minimizes the sum of the shortfalls, then the larger shortfall, with savings and score breaking remaining ties. Each target's selected configuration applies to all test datasets. The tables retain every run and report validation feasibility, including InternVL3-8B Safe-M and Safe-A, feasible in two and three of four runs, respectively.

\subsection{Data and result definitions}
\label{sec:s-protocol}
Table~\ref{tab:splits} lists per-dataset split sizes and scoring rules; the grouped splitting procedure is described in the main paper. For label generation, VQA soft scores are thresholded at 0.6 for TextVQA/VizWiz and ANLS at 0.5 for DocVQA/InfoVQA. Reported task scores retain partial credit.

Dataset-macro results average datasets equally. Relative score retention averages each dataset's method-to-original score ratio; three-target means then average the target-specific results.

\end{document}